%% file: main.tex
\pdfoutput=1

\documentclass[11pt]{article}

\usepackage{ACL2023}

\usepackage{times}
\usepackage{latexsym}

\usepackage[T1]{fontenc}

\usepackage[utf8]{inputenc}

\usepackage{microtype}

\usepackage{inconsolata}

\usepackage{multirow}
\usepackage[pdftex]{graphicx}

\usepackage{colortbl}
\usepackage{import}

\usepackage{booktabs}
\usepackage{siunitx}
\usepackage{tikz}

\usepackage{amssymb}
\usepackage{pifont}
\newcommand{\cmark}{\ding{51}}%

\usepackage{hyperref}
\usepackage{xstring}

\usepackage{color}
\usepackage[final]{changes}

\title{A Review of Datasets for Aspect-based Sentiment Analysis}

\author{Siva Uday Sampreeth Chebolu\textsuperscript{1}, Franck Dernoncourt\textsuperscript{2}, \\
{\bf Nedim Lipka\textsuperscript{3}},  
{\bf Thamar Solorio\textsuperscript{1,4}}  \\
        \textsuperscript{1}University of Houston, 
        4800 Calhoun Rd, Houston, TX, 77004, USA \\ 
        \textsuperscript{2}Adobe Research, 
        801 N 34th St, Seattle, WA 98103, USA  \\
        \textsuperscript{3}Adobe Research, 
        321 Park Ave, San Jose, CA, 95110, USA  \\
         \textsuperscript{4}MBZUAI, 
        Masdar City, Abu Dhabi, UAE  \\
        \textsuperscript{1}\texttt{\{sivauday.sampreeth8,thamar.solorio\}@gmail.com} \\ \textsuperscript{2}\texttt{franck.dernoncourt@gmail.com}, 
        \textsuperscript{3}\texttt{lipka@adobe.com}
        }

\begin{document}
\maketitle

\begin{abstract}
\input{Sections/0-abstract}
\end{abstract}

\section{Introduction}
\input{Sections/1-Introduction}
\label{Sec: Introduction}

\section{Tasks and Datasets Overview}
\input{Sections/2-Tasks-Datasets-Overview}

\label{Sec: Datasets Overview}

\section{Discussion and Future Directions}
\input{Sections/3-Discussion}
\label{Sec: Discussion and Future Directions}

\section{Conclusion}
\input{Sections/4-Conclusion}
\label{Sec: Conclusion}


\section{Limitations}
\input{Sections/6-Limitations.tex}
\label{Sec: Limitations}

\section*{Acknowledgements}
\input{Sections/7-Acknowledgements}

\bibliographystyle{acl_natbib}
\bibliography{main}

\appendix

\section{Appendix}
\input{Sections/5-Appendix}

\label{Sec: Appendix}

\end{document}

%% file: Sections/0-abstract.tex
Aspect-based sentiment analysis (ABSA) is a natural language processing problem that analyzes user-generated reviews to determine a) the target entity being reviewed, b) the high-level aspect to which it belongs, 
and c) the sentiment expressed toward the targets and the aspects. 
Numerous yet scattered corpora for ABSA make it difficult for researchers to identify corpora best suited for a specific ABSA subtask quickly. 
This study presents a database of corpora that can be used to train and evaluate autonomous ABSA systems. 
Additionally, we provide an overview of the major corpora for ABSA and its subtasks and highlight several features that researchers should consider when selecting a corpus. 
Finally, we discuss the advantages and disadvantages of existing dataset collection approaches and make recommendations for future corpora creation.
This survey examines 98 publicly available ABSA datasets covering over 25 domains, including 77 English and 21 other languages datasets (\url{https://github.com/RiTUAL-UH/ABSA-Datasets-Info}). 

%% file: Sections/1-introduction.tex



Consumers, product makers, and service providers benefit differently from user-generated reviews on e-commerce platforms.
Reading about previous customer experiences can assist future customers in making informed decisions. At the same time, the characteristics that elicit user feedback may help manufacturers and merchants develop measures to enhance customer satisfaction. 
Furthermore, as the data grows daily at a rapid pace, there is a need to recognize and extract sentiment or opinion from text reviews automatically.
Opinion mining or Sentiment Analysis \cite{sentiment-analysis-1,sentiment-analysis-2} is a technology that combines computational linguistics and natural language processing to extract such opinions. 

Aspects can be a feature, a trait, or a behavior of a product or an entity, like the atmosphere of a restaurant, the performance of a laptop, the display of a phone, and so on.
The sentiment analysis focused on a finer degree, namely, aspect-based sentiment analysis (ABSA) \cite{Hu-2004}, determines the sentiment for each entity as well as its aspects \cite{tip-of-the-iceberg}. 
Many systems, metrics, and subtasks are created along with various corpora to solve the task. 
The goal of ABSA is to extract four elements: 1) the \textit{target} entity of an opinion, 2) the high-level \textit{aspect} the entity belongs to, 3) the actual \textit{opinion} phrase, and 4) finally give a \textit{sentiment} polarity to the specific target-aspect-opinion triple. 
For example, in the review ``\textit{The pizza is very tasty.}'', \textit{pizza} is the target entity, \textit{FOOD} is the aspect category, \textit{very tasty} is the opinion phrase, and the sentiment polarity for \textit{pizza-FOOD-very tasty} is \textit{positive}. 
The subtasks get their names from the subset of identified elements in that study. 

Although there has been significant research on ABSA in the last two decades, it has become more popular after its formal introduction as a task in the SemEval-2014. SemEval-2015 consolidated its subtasks into a single framework in which all detected elements of expressed opinions (i.e., aspects, opinion and target expressions, and sentiment polarities) comply with a set of criteria and are related via sentence-level tuples. 
However, a user may be interested in the text's overall rating on a particular aspect. These ratings may be used to calculate the average sentiment for each aspect based on several sentences of a single review. 
Thus, in addition to sentence-level ABSA annotations, SemEval-2016 Task 5 included text-level ABSA
, showing that ABSA can be performed at 1) Sentence-level and 2) Review-level \cite{ACD-survey-Sampreeth}.

Given the wide range of ABSA subtasks and techniques, researchers may find it challenging to establish which corpora are optimal for a specific research task.
We want to solve this difficulty by providing an overview of available corpora and evaluating their applicability for fundamental ABSA tasks.
Specifically, this research aims to review and summarize the literature on collecting text and categorical values for ABSA elements, explain what has been learned to date, and give recommendations for constructing future datasets.

\subsection{What is different from Previous Surveys?}
\label{SubSec: Prev-Surv-Data-Crit}
    The primary difference between this survey and previous ones on ABSA \cite{Survey-ACD-SE14-15-Laskari2016,Survey-AE-Schouten2016,Survey-AE-Suresh2016,Survey-ACD-SE2016-Sethi2017,Survey-ACD-1-Sabeeh2018,Survey-ACD-DL-Do2019,Survey-ACD-1-Ahmet2020,Survey-AE-Nazir2020,ABSA-2021-Survey} is that the previous work primarily focusses on the tasks, conduct a critical analysis of the techniques, and offer ideas and future directions for enhancing the performance of the tasks and addressing unresolved issues.
    In contrast, this research aims to review and summarize the literature on collecting text reviews and categorical values for ABSA elements, explain what has been learned to date, and give recommendations for constructing future datasets. 
    Consequently, this survey complements the previous and more current ABSA surveys and critical retrospectives \cite{tip-of-the-iceberg} that focus on definitions, methodology, and evaluations.

\subsection{Contributions and Organization}
\label{SubSec: Contributions}
    We review 98 publicly available ABSA datasets in this survey that cover more than 25 domains, with 76 English and 22 other language datasets that help solve 12 different subtasks. 
    The scope of this paper covers all the datasets  specific to Aspect-based Sentiment Analysis rather than general sentiment analysis.
    We provide an overview of existing sub-tasks and current datasets, followed by a live version of the tables as a website allowing community additions. 
    Following that, we look at what can be learned from current data collection approaches and provide a few suggestions for future ABSA datasets. 
    We emphasize a few aspects including, the opinion phrase annotation and moving towards review-level ABSA from sentence-level, in the discussion section, that are particularly essential to the present ABSA research. 
        

%% file: Sections/2-Tasks-Datasets-Overview.tex
This section will discuss the various tasks and subtasks associated with ABSA and the different datasets that help solve one or more of its subtasks independently or jointly. 
    
    \subsection{Tasks Overview}
    \label{SubSec: Tasks Overview}
        ABSA comprises two sub-problems: 1) aspect extraction (for example, sushi, pasta, and well-behaved staff) and 2) identifying the polarity toward each aspect.
        Aspect extraction has two sub-tasks: a) extracting aspect terms/targets and b) categorizing/normalizing the extracted aspect terms into aspect categories.
        In turn, polarity detection has three subtasks: a) extract the opinion-oriented expression, b) determine the polarity of each category or each aspect word, and c) determine the joint polarity for aspect terms/targets and aspect categories. 
        For example, we have a positive sentiment polarity for the aspect terms \textit{value, dumplings, sushi}, and \textit{service} and the respective aspect categories \textit{Price, Food, Food}, and \textit{Service} for the review: 
        ``Highly recommend this as great value for excellent dumplings, sushi, and service''. The opinion phrases that are useful in determining the polarity are \textit{great}, and \textit{excellent}. 
        Therefore, the four main elements that we can identify from a given data for different ABSA tasks and sub-tasks 
        are 1) aspect terms/targets, 2) aspect categories, 3) opinion phrases, and 4) sentiment polarity. 

        Certain terms in Aspect-Based Sentiment Analysis (ABSA) research bear multiple labels, often used interchangeably across studies. 
        Aspect terms are often referred to as targets and opinion target expressions. Alternative usages of aspect categories can be categories and entity-attribute pairs. Opinion expressions are a span of words/tokens in a text that provide a sentiment orientation towards an aspect term or target. These can be seen as opinion phrases, opinion terms, opinion words, sentiment phrases, sentiment words, and similar vocabulary. The term \textit{aspect} can be ambiguous, as some researchers use it to denote aspect terms, while others use it for aspect categories. The exact meaning often hinges on the specific task being tackled and the elements emphasized within that task.

        \begin{table*}[!t]
            \texttt{Restaurant Review  (sentence)}: The \textbf{\textit{\color{blue}{pasta}}} was \textbf{\textit{\color{red}{very yummy}}} but the \textbf{\textit{\color{blue}{place}}} has some \textbf{\textit{\color{red}{weird smell}}}.\\ \texttt{List of Aspect Categories in the dataset:} \{\textit{Food, Ambience, Service, Price, General}\}
            \centering
            \rowcolors{2}{gray!25}{white}
            \resizebox{0.9\linewidth}{!}{%
            \begin{tabular}{lllllp{9cm}|}
            \rowcolor{gray!50}
            \hline
            \multicolumn{1}{c}{\textbf{Year}} &
              \multicolumn{1}{c}{\textbf{Paper}} &
              \multicolumn{1}{c}{\textbf{Common Name}} &
              \multicolumn{1}{c}{\textbf{Acr.}} &
              \multicolumn{1}{c}{\textbf{Input}} &
              \multicolumn{1}{c}{\textbf{Expected Output}} 
              \\ \hline
            2004 & \citet{Hu-2004}  & Aspect Term Extraction\textsuperscript{1} & ATE  & \multicolumn{1}{l}{\texttt{sentence}} & \multicolumn{1}{p{9cm}}{\textbf{\textit{\color{blue}{pasta, place}}}} 
            \\ 
            2004 & \citet{Hu-2004} & Aspect Term Sentiment Analysis\textsuperscript{1} & ATSA  & \multicolumn{1}{l}{\texttt{sentence}, (\textbf{\textit{\color{blue}{pasta, place}}})} & \multicolumn{1}{p{9cm}}{\texttt{positive, negative}} 
            \\ 
            2014 & \citet{14-dataset} & Aspect Category Detection\textsuperscript{1} & ACD   & \multicolumn{1}{l}{\texttt{sentence}} & \multicolumn{1}{p{9cm}}{\textit{Food, Ambience}}  
            \\ 
            2014 & \citet{14-dataset} & Aspect Category Sentiment Analysis\textsuperscript{1} & ACSA & \multicolumn{1}{l}{\texttt{sentence}, (\textit{Food, Ambience})} & \multicolumn{1}{p{9cm}}{\texttt{positive, negative}} 
            \\ 
            2015 & \citet{15-dataset} & Target Detection\textsuperscript{1} & TD & \multicolumn{1}{l}{\texttt{sentence}} & \multicolumn{1}{p{9cm}}{\textbf{\textit{\color{blue}{pasta, place}}}} 
            \\ 
            2014 & \citet{14-dataset} & Target Aspect Detection\textsuperscript{2} & TAD & \multicolumn{1}{l}{\texttt{sentence}} & \multicolumn{1}{p{9cm}}{(\textbf{\textit{\color{blue}{pasta}}}, \textit{Food}), (\textbf{\textit{\color{blue}{place}}}, \textit{Ambience})} 
            \\ 
            2018 & \citet{schmitt-ASD} & Aspect Sentiment Joint Detection\textsuperscript{2} & ASD & \multicolumn{1}{l}{\texttt{sentence}} & \multicolumn{1}{p{9cm}}{(\textit{Food}, \texttt{positive}), (\textit{Ambience}, \texttt{negative})} 
            \\ 
            2019 & \citet{Li-unified-TSD} & Target Sentiment Joint Detection\textsuperscript{2} & TSD & \multicolumn{1}{l}{\texttt{sentence}} & \multicolumn{1}{p{9cm}}{(\textbf{\textit{\color{blue}{pasta}}}, \texttt{positive}), (\textbf{\textit{\color{blue}{place}}}, \texttt{negative})} 
            \\ 
            2019 & \citet{ASTE-dataset} & Aspect Sentiment Triplet Extraction\textsuperscript{3} & ASTE & \multicolumn{1}{l}{\texttt{sentence}} & \multicolumn{1}{p{9cm}}{(\textbf{\textit{\color{blue}{pasta}}}, \textbf{\textit{\color{red}{very yummy}}}, \texttt{positive}), (\textbf{\textit{\color{blue}{place}}}, \textbf{\textit{\color{red}{weird smell}}}, \texttt{negative})} 
            \\ 
            2020 & \citet{TargetAspectSentimentJD} & Target Aspect Sentiment Detection\textsuperscript{3} & TASD & \multicolumn{1}{l}{\texttt{sentence}} & \multicolumn{1}{p{9cm}}{(\textbf{\textit{\color{blue}{pasta}}}, \textit{Food}, \texttt{positive}), 
            (\textbf{\textit{\color{blue}{place}}}, \textit{Ambience}, \texttt{negative})} 
            \\ 
            2020 & \citet{TOWE-dataset} & Target Opinion Word Extraction\textsuperscript{2} & TOWE & \multicolumn{1}{l}{\texttt{sentence}} & \multicolumn{1}{p{9cm}}{(\textbf{\textit{\color{blue}{pasta}}}, \textbf{\textit{\color{red}{very yummy}}}), (\textbf{\textit{\color{blue}{place}}}, \textbf{\textit{\color{red}{weird smell}}})} 
            \\ 
            2021 & \citet{ASOTE} & Aspect-Sentiment-Opinion Triplet Extraction\textsuperscript{3} & ASOTE & \multicolumn{1}{l}{\texttt{sentence}} & \multicolumn{1}{p{9cm}}{(\textbf{\textit{\color{blue}{pasta}}}, \texttt{positive}, \textit{\textbf{\color{red}{very yummy}}}), 
            (\textbf{\textit{\color{blue}{place}}}, \texttt{negative}, \textit{\textbf{\color{red}{weird smell}}})} 
            \\ 
            2021 & \citet{ACOS} & Aspect-Category-Opinion-Sentiment\textsuperscript{4}\textsuperscript{$\pm\pm$} & ACOS & \multicolumn{1}{l}{\texttt{sentence}} & \multicolumn{1}{p{9cm}}{(\textbf{\textit{\color{blue}{pasta}}}, \textit{Food}, \texttt{positive}, \textit{\textbf{\color{red}{very yummy}}}), 
            (\textbf{\textit{\color{blue}{place}}}, \textit{Ambience}, \texttt{negative}, \textit{\textbf{\color{red}{weird smell}}})} 
            \\ 
            2021 & \citet{ASQP} & Aspect Sentiment Quad Prediction\textsuperscript{4}\textsuperscript{$\pm\pm$} & ASQP & \multicolumn{1}{l}{\texttt{sentence}} & \multicolumn{1}{p{9cm}}{(\textbf{\textit{\color{blue}{pasta}}}, \textit{Food}, \texttt{positive}, \textit{\textbf{\color{red}{very yummy}}}), 
            (\textbf{\textit{\color{blue}{place}}}, \textit{Ambience}, \texttt{negative}, \textit{\textbf{\color{red}{weird smell}}})} 
            \\ 
            \hline
            \end{tabular}%

            }
            \caption{Common sub-tasks of ABSA and their relation with the identified elements Aspect Categories, Aspect Terms (a.k.a Targets), Opinion Phrases, and Sentiment Polarity. 
            {\texttt{sentence}}: Review Sentence. 
            Acr.: Acronym for the Common Name. 
            \textsuperscript{1}: single outcome sub-task, \textsuperscript{2,} \textsuperscript{3,} \textsuperscript{4}: joint outcome sub-tasks. 
            \textsuperscript{$\pm\pm$}: These tasks are similar but have been defined separately by different authors.
            }
            \label{tab: Tasks}
            \end{table*}

        We present an overview of all the sub-tasks that stemmed from ABSA in Table \ref{tab: Tasks}. 
        We provide the year and the paper in which the sub-task was first introduced along with the task name (as both acronym and full common name), the inputs of each task (as \textit{Input}), and their identified subset of elements (as \textit{Expected Output}) as the columns. 
        For example, Aspect-Category Sentiment Analysis (ACSA) aims to identify the polarity of a given aspect category. 
        However, the Target-Aspect-Sentiment Detection (TASD) task jointly identifies the targets, aspect categories, and the polarity expressed towards the target-category pair. 
        The last four rows in the Table are recently created tasks that include identifying opinion phrases in the given text that align with the sentiment polarity towards targets or aspects. 

            
    \subsection{Task Challenges}
    \label{SubSec: Task Challenges}
        There are a few major challenges for ABSA and its sub-tasks. Firstly, each of the elements described above is not independent but rather depends on other elements' detection. 
        For example, aspect term extraction and aspect category detection tasks can be used in tandem to find terms and categories in a review \cite{TargetAspectSentimentJD,14-16-MTNA}. 
        The aspect term extractor may extract related aspect terms and vice versa if it knows which aspect categories a review belongs to.
        In the review, ``\textit{However, it's the service that leaves a sour taste in my mouth}.'' The term \textit{service} is explicitly used, indicating the aspect \textit{service}.
        If the aspect term extractor is aware of the aspect category \textit{Service}, it gives the word \textit{service} in the review greater weight.
        Similarly, if the term \textit{service} is given higher weight in the review, the aspect category detector can identify the \textit{Service} category easily. 
        Also, we need to detect the implicit phrase ``sour taste in my mouth'' as a sentiment indicator to know that the review conveys a negative sentiment polarity towards \textit{service}.
        This phrase is an idiom with a negative connotation for service. The literal meaning should not be considered in this situation because the criticism is not directed toward any of the restaurant's food or beverages. There is some dedicated research on the implicit aspect and its sentiment detection \cite{Implicit-aspect}, which could be leveraged to improve the overall detection performance. 
        
        Another issue is ABSA's relevance and reliance on several other NLP tasks.
        It is worth noting that not every entity described in a text is an aspect.
        Entities that are the subject of an opinion are referred to as aspects.
        We require a sophisticated NER system to identify the names of foods, beverages, restaurants, computers, processors, hotels, and other items that may be possible targets/aspect-terms in the provided review sentence.
        To address the common opinion issue, we must find opinion phrases in the supplied review and link them to all of the proper entities. 
        This issue is closely connected to the NLP Entity-Linking problem \cite{entity-linking-2013,entity-linking-2018}.
        \textit{excellent} is used for both \textit{food} and \textit{margaritas} in the review ``\textit{the food was excellent, the margaritas too.}''
        On the other hand, there is no explicit reference to a target entity in this review ``\textit{creative and tasty but pricey},'' thus it should be assumed that the opinion is conveyed on \textit{FOOD}.
        However, we must conclude that the aspect is \textit{RESTAURANT} rather than \textit{FOOD} in the restaurant review ``\textit{Good and affordable.}''
        To address these issues, we must model the ABSA problem jointly with the related sub-problems such as aspect-term identification (NER), polarity and opinion detection (opinion phrases), syntactic simplification \cite{syntactic-simp-2006,syntactic-simp-scarton-musst} to get separate sentences for each opinion-entity pair to solve the common-opinion problem.   
        
    \subsection{Datasets Overview}
    \label{SubSec: Datasets Overview}
            All the publicly available 
            datasets for ABSA are presented in Tables \ref{tab: Datasets-no-ac} and \ref{tab: Datasets-with-ac}. 
            We provide the paper introducing the dataset, with its citation, dataset's source and domain, number of reviews and the respective number of sentences, and other statistics such as the number of sentences that are annotated with positive, negative and neutral sentiment polarity for the aspect terms/targets or aspect categories. 
            In Table \ref{tab: Dataset-Task}, we show which ABSA sub tasks from Table \ref{tab: Tasks} could be evaluated using the datasets from Table \ref{tab: Datasets-no-ac} and \ref{tab: Datasets-with-ac}. 
            
            The SemEval challenge datasets and the recently published SentiHood and MAMS corpora are the most extensively used corpora for aspect-based sentiment analysis.
            The SemEval corpora were made public as part of a shared work held during the International Workshop on Semantic Evaluation, held annually from 2014 to 2016. The datasets are described in full in \cite{14-dataset,15-dataset,16-dataset}.
            Historically, ABSA was primarily concerned with aspect term extraction and sentiment analysis \cite{Hu-2004}. Before 2014, there has been very little research into aspect category detection and sentiment analysis.
            However, as ACD was formally presented at SemEval-2014, a slew of new challenges arose.
            ABSA has witnessed positive outcomes across all tasks thanks to the emergence of artificial neural networks.
            
            Despite the popularity of SemEval datasets for this work, most sentences only include one or many aspects with the same sentiment polarity, reducing the ABSA task to sentence-level sentiment analysis.
            \cite{MAMS-dataset} published a new large-scale Multi-Aspect Multi-Sentiment (MAMS) dataset, in which each phrase has at least two independent aspects with different sentiment polarity. 
            \added{
                Although \citet{MAMS-dataset} claimed that each sentence has more than one aspect-sentiment tuple, the approach they followed is not realistic. When there is only one opinion tuple in a sentence, they introduce either a ``miscellaneous'' category or another category with neutral sentiment as a second opinion tuple that does not have an opinion in the review. For instance, in the following review, \textit{I like the smaller portion size for dinner.}, there is only one opinion, which is about the food's portion size. However, the actual annotation has two opinion tuples: one is for the food, and the other is a neutral opinion on the restaurant's \textit{miscellaneous} aspect category. We do not dispute the legitimacy of this strategy, but we do not find it practical in the real world. 
            }
            Another drawback with the SemEval corpora is that they contain reviews about a single \textit{target entity}, such as a laptop or restaurant. 
            To overcome this, \cite{sentihood-dataset} created the SentiHood dataset to identify the sentiment towards each aspect of one or more entities. 
            
            As discussed previously, an opinion phrase is critical in determining the sentiment polarity towards an aspect or a target and, sometimes, determining to which target/aspect that opinion belongs. 
            \citet{TOWE-dataset,ASTE-dataset} modified the SemEval datasets to account for the missing annotation of opinion phrases that lead to a specific sentiment polarity for the target, aspect term, or aspect category. 
            However, this resulted in a few instances, coercing them to merge all the reviews from SemEval 2014 to 2016 into a single dataset.

        \begin{table*}[!h]
        \centering
        \rowcolors{2}{gray!25}{white}
        \resizebox{0.9\textwidth}{!}{%
        \begin{tabular}{p{7cm}cccccccccccc}
            \rowcolor{gray!50}
                \hline
                \multicolumn{1}{c}{\textbf{Dataset Paper}} &
                  \textbf{ATE} &
                  \textbf{ATSA} &
                  \textbf{ACD} &
                  \textbf{ACSA} &
                  \textbf{TD} &
                  \textbf{TSD} &
                  \textbf{ASD} &
                  \textbf{TAD} &
                  \textbf{TASD} &
                  \textbf{ASTE} &
                  \textbf{TOWE} &
                  \textbf{QUAD.Ex} \\ \hline
                Customer Reviews \cite{Bing-Liu-Products1} &
                  \cmark &
                  \cmark &
                  - &
                  - &
                  - &
                  - &
                  - &
                  - &
                  - &
                  - &
                  - &
                                    - \\ 
                Customer Reviews \cite{Bing-Liu-Products2} &
                  \cmark &
                  \cmark &
                  - &
                  - &
                  - &
                  - &
                  - &
                  - &
                  - &
                  - &
                  - &
                                    - \\ 
                JDPA \cite{JDPA-Sentiment-Corpus} &
                  - &
                  - &
                  - &
                  - &
                  \cmark &
                  \cmark &
                  - &
                  - &
                  - &
                  - &
                  - &
                                    - \\ 
                Darmstadt Service \cite{Darmstadt-Service-Review-Corpus} &
                  - &
                  - &
                  - &
                  - &
                  \cmark &
                  \cmark &
                  - &
                  - &
                  - &
                  - &
                  - &
                                    - \\ 
                TripAdvisor Hotels \cite{TripAvisor-AmazonMp3-dataset1} &
                  - &
                  - &
                  \cmark &
                  \cmark &
                  - &
                  - &
                  \cmark &
                  - &
                  - &
                  - &
                  - &
                                    - \\ 
                Czech Restaurants \cite{CZech-ABSA} &
                  \cmark &
                  \cmark &
                  \cmark &
                  \cmark &
                  - &
                  - &
                  - &
                  - &
                  - &
                  - &
                  - &
                                    - \\ 
                SE-14 Restaurants \cite{14-dataset} &
                  \cmark &
                  \cmark &
                  \cmark &
                  \cmark &
                  - &
                  - &
                  - &
                  - &
                  - &
                  - &
                  - &
                                    - \\ 
                SE-14 Laptops \cite{14-dataset} &
                  \cmark &
                  \cmark &
                  - &
                  - &
                  - &
                  - &
                  - &
                  - &
                  - &
                  - &
                  - &
                                    - \\ 
                Twitter Comments \cite{Target-Dependent-Twitter-Sentiment} &
                  - &
                  - &
                  - &
                  - &
                  \cmark &
                  \cmark &
                  - &
                  - &
                  - &
                  - &
                  - &
                                    - \\ 
                Customer Reviews \cite{Bing-Liu-Products3} &
                  \cmark &
                  \cmark &
                  - &
                  - &
                  - &
                  - &
                  - &
                  - &
                  - &
                  - &
                  - &
                                    - \\ 
                HAAD \cite{Arabic-book} &
                  - &
                  - &
                  \cmark &
                  \cmark &
                  \cmark &
                  \cmark &
                  \cmark &
                  - &
                  \cmark &
                  - &
                  - &
                                    - \\ 
                SE-15 Restaurants \cite{15-dataset} &
                  - &
                  - &
                  \cmark &
                  \cmark &
                  \cmark &
                  \cmark &
                  \cmark &
                  - &
                  \cmark &
                  - &
                  - &
                                    - \\ 
                SE-15 Laptops \cite{15-dataset} &
                  - &
                  - &
                  \cmark &
                  \cmark &
                  - &
                  - &
                  \cmark &
                  - &
                  - &
                  - &
                  - &
                                    - \\ 
                SE-16 Rest \& Hotels \cite{16-dataset} &
                  - &
                  - &
                  \cmark &
                  \cmark &
                  \cmark &
                  \cmark &
                  \cmark &
                  - &
                  \cmark &
                  - &
                  - &
                                    - \\ 
                SE-16 Telecom \cite{16-dataset} &
                  - &
                  - &
                  \cmark &
                  \cmark &
                  \cmark &
                  \cmark &
                  \cmark &
                  - &
                  \cmark &
                  - &
                  - &
                                    - \\ 
                SE-16 Laptops \cite{16-dataset} &
                  - &
                  - &
                  \cmark &
                  \cmark &
                  - &
                  - &
                  \cmark &
                  - &
                  - &
                  - &
                  - &
                                    - \\ 
                SE-16 Mob.Phns.\cite{16-dataset} &
                  - &
                  - &
                  \cmark &
                  \cmark &
                  - &
                  - &
                  \cmark &
                  - &
                  - &
                  - &
                  - &
                                    - \\ 
                SE-16 Dig.Cam. \cite{16-dataset} &
                  - &
                  - &
                  \cmark &
                  \cmark &
                  - &
                  - &
                  \cmark &
                  - &
                  - &
                  - &
                  - &
                                    - \\ 
                SentiHood \cite{sentihood-dataset} &
                  - &
                  - &
                  \cmark &
                  \cmark &
                  \cmark &
                  \cmark &
                  \cmark &
                  - &
                  \cmark &
                  - &
                  - &
                                    - \\ 
              GermEval-2017 \cite{germevaltask2017} &
                  - &
                  - &
                  \cmark &
                  \cmark &
                  \cmark &
                  \cmark &
                  \cmark &
                  - &
                  \cmark &
                  - &
                  - &
                                    - \\ 
                BeerAdvocate, TripAdvisor \cite{BeerAdvocate-TripAdvisor-dataset} &
                  - &
                  - &
                  \cmark &
                  \cmark &
                  - &
                  - &
                  \cmark &
                  - &
                  - &
                  - &
                  - &
                                    - \\ 
                ABSITA-2018 \cite{ABSITA-2018} &
                  - &
                  - &
                  \cmark &
                  \cmark &
                  - &
                  - &
                  \cmark &
                  - &
                  - &
                  - &
                  - &
                                    - \\ 
                FiQA \cite{FiQA-dataset} &
                  - &
                  - &
                  \cmark &
                  \cmark &
                  \cmark &
                  \cmark &
                  \cmark &
                  - &
                  \cmark &
                  - &
                  - &
                                    - \\ 
                Ba-Re-Cr \cite{Bangla} &
                  - &
                  - &
                  \cmark &
                  \cmark &
                  - &
                  - &
                  \cmark &
                  - &
                  - &
                  - &
                  - &
                                    - \\ 
                MAMS \cite{MAMS-dataset} &
                  \cmark &
                  \cmark &
                  \cmark &
                  \cmark &
                  - &
                  - &
                  \cmark &
                  - &
                  - &
                  - &
                  - &
                                    - \\ 
                TOWE \cite{TOWE-dataset} &
                  - &
                  - &
                  - &
                  - &
                  \cmark &
                  - &
                  - &
                  - &
                  - &
                  - &
                  \cmark &
                                    - \\ 
                Telugu Movies \cite{telugu-dataset} &
                  \cmark &
                  \cmark &
                  \cmark &
                  \cmark &
                  - &
                  - &
                  \cmark &
                  \cmark &
                  - &
                  - &
                  - &
                                    - \\ 
                ASTE \cite{ASTE-dataset} &
                  - &
                  - &
                  - &
                  - &
                  \cmark &
                  \cmark &
                  - &
                  - &
                  - &
                  \cmark &
                  \cmark &
                                    - \\ 
                ASOTE \cite{ASOTE} &
                  - &
                  - &
                  - &
                  - &
                  \cmark &
                  \cmark &
                  - &
                  - &
                  - &
                  \cmark &
                  \cmark &
                                    - \\ 
                ABSITA-2020 \cite{ABSITA-2020} &
                  \cmark &
                  \cmark &
                  - &
                  - &
                  - &
                  - &
                  - &
                  - &
                  - &
                  - &
                  - &
                                    - \\ 
                NewsMTSC \cite{NewsMTSC} &
                  \cmark &
                  \cmark &
                  - &
                  - &
                  - &
                  - &
                  - &
                  - &
                  - &
                  - &
                  - &
                                    - \\ 
                ASAP \cite{chinese-rest-asap-2021} &
                  - &
                  - &
                  \cmark &
                  \cmark &
                  - &
                  - &
                  \cmark &
                  - &
                  - &
                  - &
                  - &
                                    - \\ 
                ASQP \cite{ASQP} &
                  \cmark &
                  \cmark &
                  \cmark &
                  \cmark &
                  \cmark &
                  \cmark &
                  \cmark &
                  \cmark &
                  \cmark &
                  \cmark &
                  \cmark &
                                    \cmark \\ 
                ACOS \cite{ACOS} &
                  \cmark &
                  \cmark &
                  \cmark &
                  \cmark &
                  \cmark &
                  \cmark &
                  \cmark &
                  \cmark &
                  \cmark &
                  \cmark &
                  \cmark &
                                    \cmark \\
                DM-ASTE \cite{DM-ASTE} &
                  - &
                  - &
                  - &
                  - &
                  \cmark &
                  \cmark &
                  - &
                  - &
                  - &
                  \cmark &
                  \cmark &
                                    - \\
                DE-ASTE \cite{Dom-Exp-ASTE} &
                  - &
                  - &
                  - &
                  - &
                  \cmark &
                  \cmark &
                  - &
                  - &
                  - &
                  \cmark &
                  \cmark &
                                    - \\ 
                MEMD-ABSA \cite{DM-ASTE} &
                  \cmark &
                  \cmark &
                  \cmark &
                  \cmark &
                  \cmark &
                  \cmark &
                  \cmark &
                  \cmark &
                  \cmark &
                  \cmark &
                  \cmark &
                                    \cmark \\\hline

                \end{tabular}%
        }
        \caption{Related subtasks of ABSA for each Dataset. QUAD.Ex: Quadruple Extraction. Ba-Re-Cr: Bangla Restaurants and Cricket Dataset. Note: The acronyms of the subtasks in the column names are according to the \textit{Acr.} column in Table \ref{tab: Tasks}. }
        \label{tab: Dataset-Task}
        \end{table*}
            
    
    \subsection{Annotation Procedure and Dataset Source}
    \label{SubSec: Annotation Procedures}
            Even though researchers use various annotation methods when building ABSA datasets, we explain the most frequent method here.
            One annotator (A) initially annotates a portion of the data, which is then checked by another annotator (B) for any corrections.
            The remainder of the sentences in the dataset will be annotated by annotator A, with additional instructions based on the nature of the earlier disagreements.
            When A lacked assurance, a decision was taken in collaboration with B. When A and B differed, they and a third expert annotator came to a judgment together. 
            Another conflict resolution method was to take the vote of the majority and consider that as the correct annotation. 
            Since most of the datasets follow this procedure where one annotator annotates and the expert annotator checks for the mistakes, many of the dataset papers lack the inter-annotator agreement scores. 
            
            The SemEval-2014 (SE-14) dataset was annotated in two stages. The first stage consisted of tagging and detecting the polarity of all single and multi-word words that designated certain aspects of the target item.
            The second step involves identifying the aspect categories and polarity of the sentences.
            Most datasets that include annotations simply for aspect terms/targets and their polarity, such as the Customer Review datasets \cite{Bing-Liu-Products1,Bing-Liu-Products2,Bing-Liu-Products3}, TOWE \cite{TOWE-dataset}, ASTE \cite{ASTE-dataset}, follow the first stage of this process.
            The second stage is only implemented for datasets including aspect categories, such as the SemEval and FiQA datasets.
            
            Few datasets, such as MAMS, give distinct annotations for ATSA and ACSA tasks where there is no one-to-one correspondence between aspect words, aspect categories, and their polarities.
            The restaurant datasets from SemEval (Table \ref{tab: Datasets-no-ac} and \ref{tab: Datasets-with-ac}) are a subset of the datasets published by \citet{Ganu-CitySearch} that had only six aspect categories.
            
            A typical approach followed by researchers is to take existing datasets and annotating them for missing items for an existing subtask or propose a new subtask for ABSA from the annotations.
            The TOWE and ASTE datasets (Table \ref{tab: Datasets-no-ac}) are derived from SemEval restaurants and laptops. 
            The authors included the opinion phrase information for the existing opinion tuples to propose a new subtask. 
            The most common disagreements were noticed when annotating the multiword aspect term boundaries, aspect term vs. reference to target entity, neutral polarity ambiguity, and the problem of distinguishing aspect terms when they appear in conjunctions or disjunctions. 
            The last one was resolved using the maximal phrase as the aspect term.             
            Most of the English restaurant datasets, such as SemEval, MAMS, TOWE, and ASTE, are obtained from citysearch.com for New York restaurants. while the Laptop data were derived from laptop reviews on Amazon.com. 
            Since aspect category detection task is formmaly introduced in SemEval-2014, all preceding datasets only include annotations for aspect words and their polarity.

%% file: Sections/3-Discussion.tex
We explore several characteristics of the corpora in this section, including the formats, the need for joint datasets with opinion phrase annotations, and recommendations for future ABSA datasets.

    \subsection{Dataset Formats}
    \label{SubSec: Formats}
            The definition and format of ABSA components vary greatly   depending on the dataset's source.
            SemEval-2014, for example, published a dataset with explicit and independent aspect categories, aspect terms, and corresponding sentiment polarity.
            Because there is no one-to-one correspondence between the terms and the categories, using aspect terms and categories in a joint detection scenario is problematic.
            Therefore, one is forced to work on either the ATE and ASTE or the ACD and ACSA.
            However, in SemEval-2015 and SemEval-2016, the dataset structure is more unambiguous, establishing a link between the targets and the aspect categories. The sentiment polarity is linked to the target-aspect category combination.
            It allows the community to recognize a text's stated sentiment better using terms and categories.
            Again, in SemEval-2015 and SemEval-2016, the aspect category is divided into 1) Entity and 2) Attribute. 
            Entities can be the reviewed entity itself, such as the \textit{RESTAURANT}, a part/component of it, such as \textit{AMBIENCE}, or another relevant entity, such as \textit{DRINKS}.  
            Attributes are facets of an entity such as \textit{PRICE} or \textit{QUALITY}. 
            
            \citet{MAMS-dataset}, \citet{telugu-dataset} and a few others followed the SemEval-2014's XML format and released new datasets in the recent past. 
            However, \cite{TOWE-dataset,ASTE-dataset} modified the datasets from all three SemEval shared tasks into another format to include the opinion phrase
            \added{
                and released the datasets in an XML and NER task's BIO (beginning, inside, and outside) format. 
                }
                
            On the other hand, the SentiHood dataset used  a \textit{JSON} format to provide the annotations for targets, aspects, and sentiments. But the definition of aspect category in the SentiHood dataset is the combination of the target and the aspect, leading to identifying ACD and ACSA tasks from Table \ref{tab: Tasks}. 
            As mentioned in Section \ref{SubSec: Datasets Overview}, aspect categories are formally introduced in SemeEval-2014. All the prior datasets only annotate the aspect terms and their sentiment polarity to solve the ATE and the ATSA tasks. 
            Therefore, for more robust ABSA systems, we urge that the community use an already established structure and criteria for future datasets rather than introducing a new format or structure. 
            The standardization of the annotation format would also benefit benchmarking and updating existing models without any adjustments to the  architecture.

    \subsection{Opinion Phrases and Datasets Merging}
    \label{SubSec: Merging Datasets}
            As previously explained in Section \ref{SubSec: Task Challenges}, we must identify the opinion words in a given text to determine the sentiment polarity and the entities on which the opinion is conveyed, i.e., the aspect terms. 
            It is evident from Table \ref{tab: Dataset-Task} that most recent tasks, such as the ASTE, TOWE, and ASQP, annotated the opinion words in the current SemEval shared task datasets to enhance the ABSA task. 
            
            In the original datasets of the SemEval challenge, the opinion targets (aspect terms) are annotated, but the opinion words and their correspondence with targets are not provided. 
            In addition, most of the available benchmark corpora are small. This gives the opportunity to combine or merge datasets with similar characteristics. For instance, \citet{TOWE-dataset} annotated the corresponding opinion words for the annotated targets. 
            The sentences without targets or with implicit opinion expressions are not included. The original ASTE dataset does not contain cases where a single opinion span is associated with multiple targets. Consequently, \citet{ASTE-dataset} refined the dataset with these additional missing triplets and expanded the corpora. 
            
            The community could focus on merging the existing datasets to obtain better quality corpora with increased sizes. Furthermore, researchers could annotate for the opinion words missing in most of the current datasets, which could greatly improve the overall performance of ABSA. 

   \begin{table*}[!h]
        \texttt{Review}: I picked the \textbf{\textit{\color{blue}{asparagus}}}, which was \textbf{\textit{\color{red}{incredible}}}. It was \textbf{\textit{\color{red}{steamed and tossed with garlic}}}. The \textbf{\textit{\color{blue}{steak}}} was nice and juicy. It's \textbf{\textit{\color{red}{served with either a peppercorn sauce or red wine reduction}}}. The \textbf{\textit{\color{blue}{service from the staff}}} was \textbf{\textit{\color{red}{extremely attentive and very friendly}}}. It was the highlight of our dinner.\\
        \centering
        \resizebox{0.8\textwidth}{!}{%
        \begin{tabular}{llll}
        \hline
        \multicolumn{1}{c}{ \textbf{Target}} &
          \multicolumn{1}{c}{\textbf{Aspect Cat.}} &
          \multicolumn{1}{c}{\textbf{Opinion Expr}} &
          \multicolumn{1}{c}{\textbf{Sent.}} \\ \hline
        {\color[HTML]{0000FF} asparagus} &
          {\color[HTML]{6666FF} FOOD\#QLTY} &
          {\color[HTML]{707003} incredible} &
          {\color[HTML]{CC0000} Positive} \\ 
        \cellcolor[HTML]{FFB6C1}{\color[HTML]{0000FF} NULL} &
          {\color[HTML]{6666FF} FOOD\#STY\_OP} &
          {\color[HTML]{707003} steamed and tossed with garlic} &
          {\color[HTML]{CC0000} Positive} \\ 
        {\color[HTML]{0000FF} steak} &
          {\color[HTML]{6666FF} FOOD\#QLTY} &
          {\color[HTML]{707003} nice and juicy} &
          {\color[HTML]{CC0000} Positive} \\ 
        \cellcolor[HTML]{FFB6C1}{\color[HTML]{0000FF} NULL} &
          {\color[HTML]{6666FF} FOOD\#STY\_OP} &
          {\color[HTML]{707003} served with either a peppercorn sauce or red wine reduction} &
          {\color[HTML]{CC0000} Positive} \\ 
        {\color[HTML]{0000FF} service from the staff} &
          {\color[HTML]{6666FF} SERVICE\#GEN} &
          {\color[HTML]{707003} extremely attentive and very friendly} &
          {\color[HTML]{CC0000} Positive} \\ 
        \cellcolor[HTML]{FFB6C1}{\color[HTML]{0000FF} NULL} &
          {\color[HTML]{6666FF} \begin{tabular}[c]{@{}l@{}}FOOD\#QLTY (OR)\\ AMBIANCE\#GEN\end{tabular}} &
          {\color[HTML]{707003} highlight} &
          {\color[HTML]{CC0000} Positive} \\ \hline
        \end{tabular}%
        }
        \caption{Result of applying Sentence-Level methods to full-review to detect Target, Aspect, Opinion, and Sentiment. 
        }
        \label{tab: OATS-Sentence-Level}
    \end{table*}

    \subsection{Need Large Datasets for Unified Models}
    \label{SubSec: Large Datsets}
            Recent ABSA datasets are mostly drawn from SemEval shared challenges and include additional data processing and task-specific annotations. The small number of instances (for example, hundreds of phrases) in each dataset makes it challenging to compare models with reliability, particularly Transformer-based models with millions of parameters. 
            Researchers currently evaluate a model's accuracy by averaging the results of numerous runs, but larger datasets would allow for more precise comparisons. However, more challenging datasets must still be provided to meet real-world scenarios that include reviews from many domains such as \citet{DM-ASTE,Dom-Exp-ASTE} or languages, for example, can help evaluate multi-domain and multi-lingual ABSA systems. 

            
            Recently, the unified models built using the generative frameworks \cite{T5-ABSA-sampreeth,T5-ABSA-towards-generative} yield SOTA performance on all the subtasks of ABSA by jointly solving for all the elements. 
            The advantage of these unified models is that they could solve multiple subtasks without a change in the model architecture. 
            Building more datasets similar to \citet{ASQP,MEMD-ABSA}, with annotations for all the elements, would be beneficial in developing and investigating these promising types of models.

    \subsection{Datasets in Low-Resource Languages}
    \label{SubSec: Low Resource Languages}
            It is evident from the presented tables and preceding discussion that the majority of ABSA's available datasets are in English. 
            Very few datasets are available for low-resource languages, where in most cases, just one dataset is accessible. 
            Numerous studies employed cross-lingual approaches to automatically produce new datasets for low-resource languages, or to tackle the target language using data from the source language, or for other purposes. 
            A few works first translated from the source to the target language with an off-the-shelf translation system and then aligned the labels using FastAlign or some other softwares \cite{cros-lingual-exploring-distrb,cross-lingual-instance-selection}. 
            Others use the cross-lingual word embeddings pre-trained on large parallel bilingual corpus, or utilize the multilingual Pre-trained Language Models (mPLMs) such as multilingual BERT \cite{BERT} and XLM-RoBERTa \cite{cross-lingual-unsupervised}, or the zero-shot learning \cite{cross-lingual-zero-shot}. 
            Despite these efforts, the quality of the datasets for low-resource languages are still under par. 
            It could also be attributed to the fact that the XABSA problem is relatively under-explored compared to the monolingual ABSA. Although mPLMs are widely used for various cross-lingual NLP tasks nowadays, exploring their usage in the XABSA can be tricky since language-specific knowledge plays an essential role in any ABSA task \cite{ABSA-methods-2022-survey}. 
            
            It explains the need to annotate datasets in different languages to push the boundaries of ABSA research ahead. 
            Given the fact that sentiment analysis is a very subjective task, if researchers do not compromise on getting high inter-annotator agreement scores for annotating different elements of ABSA, the quality of the datasets could be ensured with high-confidence. 
            As an initial step, taking the annotation costs into consideration, researchers can build small but high quality datasets in the low-resource languages.

    \subsection{Sentence-Level to Review-Level ABSA}
    \label{SubSec: Sent-to-Review-level}

        In real-life scenarios, reviews often contain multiple sentences with overlapping contexts (Figure \ref{fig: Multi-sent-reviews} in Appendix), making sentence-level ABSA methods less effective. These methods assign a NULL value for implicit targets and opinions within a sentence context and struggle to generalize to full reviews. In contrast, by considering the entire review, explicit targets in one sentence might be referred to implicitly in others, providing a richer and more nuanced analysis. The previous focus on sentence-level ABSA limits the applicability of these methods to real-world situations. Even tasks that included a Text-Level ABSA component, like SE16-ABSA \cite{16-dataset,ACD-survey-Sampreeth}, were aimed at summarizing opinions from individual sentences rather than capturing ABSA elements within the entire review context.
        
        Moreover, applying sentence-level ABSA to full reviews often fails to accurately capture the intended targets of pronouns or implicit references (Table \ref{tab: OATS-Sentence-Level}). For instance, the target of a pronoun like "it" will be marked as NULL (implicit) when sentences are considered in isolation. However, in the broader context of the review, "it" might refer to a specific entity mentioned in previous sentences. This lack of context makes it difficult for models to correctly assign opinions to targets.
        
        Implicit targets could be effectively handled by considering the full context of a review. Likewise, the identification of correct categories also benefits from full context analysis (Table \ref{tab: ex-OATS} in appendix). While previous benchmark datasets from SemEval competitions annotate opinions in a sentence based on review context, this approach is not guaranteed to generalize to all implicit cases, potentially leading to inconsistencies and reduced performance in system predictions. Hence, the adoption of review-level ABSA and the creation of corresponding datasets can significantly improve the effectiveness of sentiment analysis.
        
    \subsection{Inter-Annotator Agreement}
    \label{SubSec: IAA}
        The inter-annotator agreement, a key metric for evaluating the quality of a dataset, can be quantified in various ways such as the renowned Cohen's Kappa and Fleiss's Kappa scores, which are used for classification tasks. However, according to a study \cite{not-kappa}, Kappa may not be the best fit for span-extraction annotation in textual data. The limitation arises from the requirement of Kappa to compute the number of negative cases, which is unidentifiable for spans as they constitute sequences of words without a predetermined quantity of items for annotation in a text.
    
        
        Due to these limitations of Kappa metric, the F-measure, which doesn't necessitate the calculation of negative cases, is often more suitable for gauging inter-annotator agreement in span extraction annotation tasks such as target and opinion phrase extraction \cite{IAA-F1}. In particular, for datasets that encompass both classification (category and polarity) and span extraction tasks, such as the ASQP dataset, the F-measure can effectively serve as the chief method for inter-annotator agreement computation. 

    \subsection{Is ABSA only for reviews?}
    \label{SubSec: ABSA-only-Reviews}
        A significant limitation in the current landscape of ABSA datasets is the predominant focus on reviews and specific domains like restaurants and e-commerce platforms. This focus on customer reviews has been fostered by the abundance and accessibility of data in these areas. For instance, review websites and e-commerce platforms readily provide vast amounts of customer feedback data. While these applications have been successful, the over-reliance on specific data types curtails the broader applicability of ABSA and its potential to provide diverse insights across various sectors. 
        
        Diverse ABSA datasets are vital for advancing research and applications beyond the confines of reviews. By expanding datasets to encompass domains such as healthcare, education, finance, legal, and social issues, ABSA models can be trained to tackle real-world challenges and address broader problem areas. For instance, healthcare datasets could facilitate sentiment analysis of patient feedback, improving healthcare service quality. 
        ABSA can help identify investor sentiment towards different aspects of a company or its financial performance, potentially predicting stock market trends \cite{Fin-Entity-Sent-Ana,FiQA-dataset,finxabsa,Bangla-Fin-News}.
        Education datasets could uncover student sentiments towards specific aspects of the learning environment, leading to targeted improvements \cite{Arabic-Education}. 
        
        Expanding datasets to these areas would enable the development of more robust and generalizable models while enhancing decision-making processes, public opinion analysis, and customer experiences in diverse industries. 

    \subsection{Why evaluate only on a few datasets?}
    \label{SubSec: why-only-few-datasets}
        While ABSA research has numerous datasets at its disposal, the focus often falls on a select few benchmark datasets. This practice, while providing consistency and quality control, might unintentionally narrow model adaptability and heighten biases. Benchmark datasets may not cover the breadth of linguistic diversity in real-world scenarios, potentially causing models to falter with different or new data. Over-reliance on these resources can instigate model bias as models may echo the limitations of their training data, compromising performance in varied contexts. 
        
        Addressing these issues necessitates the use of an array of datasets for evaluation and the creation of a central platform for evaluating all ABSA models across available datasets. By offering researchers a unified platform for accessing diverse datasets and generating standardized metrics, we can advance understanding of model performance. This crucial step in ABSA research can stimulate more adaptive and resilient model development. The proposed platform would not only assist in overcoming the limitations of benchmark datasets but also inspire the creation of new, superior ones. While similar initiatives have been seen in the broader NLP community \cite{lince}, ABSA-specific platforms are still a necessity. This progressive move calls for the combined effort of researchers and professionals alike to pave the way for more effective and fair sentiment analysis tools.

    \subsection{Baseline Study: ABSA Task Variations Across Datasets}
        One of the main questions that comes to mind is why we need new datasets for the same task(s). 
        Let us look at a sample of datasets from Table \ref{tab: Dataset-Task} and analyze the performance of six different ABSA subtasks from Table \ref{tab: Tasks} using the unified generative paraphrasing framework \cite{ASQP}.
        The main objective of this experiment is to assess and contrast the intricacy of various datasets concerning multiple tasks. Additionally, we endeavor to illustrate the inter-dependencies among these tasks, emphasizing the imperative nature of addressing these dependencies to enhance task performance on a given dataset.
        
        We chose three types of joint tasks for our experiments: (a) tuple extraction (ASD, TSD, AOPE), (b) triplet extraction (ASTE, TASD), and (c) quadruple extraction (ASQP). 
        The emphasis was placed on tasks involving joint extraction rather than single-element extraction. The rationale behind this decision stems from findings in \citet{T5-ABSA-sampreeth}, where the authors have shown that joint models, designed to handle multiple interrelated elements simultaneously, consistently outperformed models that were specifically fine-tuned for extracting a single element. The results of these experiments can be found in Table \ref{tab: Baseline-Exp}.

        \begin{table}[!h]
            \centering
            \rowcolors{2}{gray!25}{white}
            \resizebox{0.9\columnwidth}{!}{%
            \begin{tabular}{lrrrrrr}
            \rowcolor{gray!50}
            \hline
            \textbf{Dataset} &
              \multicolumn{1}{c}{\textbf{ASD}} &
              \multicolumn{1}{c}{\textbf{TSD}} &
              \multicolumn{1}{c}{\textbf{AOPE}} &
              \multicolumn{1}{c}{\textbf{ASTE}} &
              \multicolumn{1}{c}{\textbf{TASD}} &
              \multicolumn{1}{c}{\textbf{ASQP}} \\ \hline
            ACOS\_Rest            & 77.03      & 75.01 & 71.25 & 68.77 & 71.71 & 65.71 \\
            ACOS\_Lap             & 53.15      & 74.05 & 75.21 & 74.43 & 48.21 & 57.30 \\
            DM\_ASTE\_Beauty      & \textbf{-} & 65.05 & 42.94 & 45.26 & -     & -     \\
            DM\_ASTE\_Electronics & -          & 65.18 & 43.13 & 44.76 & -     & -     \\
            DM\_ASTE\_Home        & -          & 67.85 & 43.93 & 44.44 & -     & -     \\
            DM\_ASTE\_Fashion     & -          & 66.27 & 42.70 & 44.62 & -     & -     \\
            SE-15                 & 71.10      & 69.31 & 57.38 & 62.56 & 63.06 & 46.93 \\
            SE-16                 & 76.97      & 75.60 & 65.80 & 71.70 & 71.97 & 57.93 \\ \hline
            \end{tabular}%
            }
            \caption{Baseline experimental results for six tasks from Table \ref{tab: Tasks} using Paraphrase-T5 method from \citet{ASQP}. - indicates the dataset doesn't have relevant data for that experiment. }
            \label{tab: Baseline-Exp}
        \end{table}
            
        Two combinations in these six tasks essentially affect ASQP positively or negatively: 1) ASD + TSD + AOPE and 2) ASTE + TASD. 
        For the ACOS\_Rest dataset, the performance of the ASTE and TASD task together helped the ASQP task to detect the quadruples decently. 
        On the other hand, in the ACOS\_Lap dataset, TSD, AOPE, and ASTE have significantly better performance when compared to ASD and TASD. This shows that identifying aspect categories in the laptop dataset is difficult compared to targets and opinion phrases, resulting in the low performance of the ASQP task. 
        On a similar note, the DM\_ASTE datasets have consistent performance across all the domains. However, the performance of the AOPE task is far from the TSD task, indicating that it is challenging to comprehend the complex target and opinion phrase relationship compared to the target sentiment relationship. It negatively affected the ASTE performance, a combination of TSD+AOPE tasks. 

        Datasets with higher performance in lower-level tasks don't always guarantee high performance in comprehensive tasks like ASQP. For instance, in SE-16, despite ASTE and TASD being reasonably high, ASQP was lower, pointing to complexities in merging the triplets. 
        Given the presence of opinion phrases in AOPE, ASTE, and ASQP, the clarity and diversity of opinion expressions in a dataset can be pivotal. If AOPE scores are significantly lower than TASD or ASD, it can explain why ASQP might suffer. 
        This is clearly evident in the se-15 and se-16 datasets. Furthermore, ASD and TSD scores in rest\_acos show the dataset has clear sentiment expressions for categories and targets. This clarity in individual extraction didn't fully translate to ASTE and ASQP, suggesting that extracting opinion words/phrases concurrently with targets and aspects might be challenging. 
        

        In conclusion, it is evident that each of these datasets presents distinct challenges, ranging from identifying opinion terms in the SE-15 dataset to focusing on aspect categories in the ACOS\_lap dataset. 
        Further research is required to fully harness the inter-dependencies that exist among these tasks utilizing the datasets. Additionally, it is crucial to investigate whether the sub-optimal performance observed is attributed to the intrinsic complexity of the task, the inherent ambiguity of the elements, or the inadequacies in their representation. 

%% file: Sections/4-Conclusion.tex

In this survey, we highlighted the urgent need for standardization and diversity in ABSA datasets to ensure comparability and enhance model robustness. Emphasizing the value of opinion phrases and the potential benefits of merging similar datasets, we advocated for the creation of resources in low-resource languages and shifting focus to review-level ABSA. Highlighting the importance of robust inter-annotator agreement measures, we called for an expansion of ABSA beyond reviews and addressed the limitations of evaluating ABSA methods on only a few datasets. Consequently, we proposed a common platform for ABSA evaluation to foster comprehensive and fair assessments, promoting the development of more effective sentiment analysis tools. With these initiatives, we can advance ABSA research and ensure its applicability across diverse linguistic and domain contexts.

%% file: Sections/6-Limitations.tex
While this survey endeavored to encompass a broad array of ABSA datasets, it is conceivable that some may have been unintentionally missed. Further, due to space constraints, we were unable to delve deeply into datasets from other crucial domains, including legal, healthcare, and education sectors. 
Even though we performed a few experiments to understand the interplay of datasets and tasks, we did not explore the option of combining the model predictions from ASTE + TASD and ASD + TSD + AOPE tasks, similar to an ensemble approach, to solve for ASQP. 
However, it is a non-trivial task to combine the predictions from the triplets, as target and sentiment polarity are the only common elements to merge, which may not be unique for an aspect category and opinion phrase pairs to form a quadruple. That is, the same target an sentiment polarity pair can exist for different aspect categories and different opinion phrases. 

Although we have provided accessible links for public datasets, acquisition of other datasets necessitates direct requests to the original authors or proprietors. We encourage the community's active involvement in contributing to a live version of this review to address these identified gaps. Another potential limitation pertains to the absence of a discussion on data ownership and copyrights issues related to datasets obtained via web scraping. Web scraping often contravenes the terms of service of the scraped websites, and disseminating scraped content might infringe on copyright laws, particularly if the scraped data have been substantially altered. Hence, it is crucial to exercise due diligence in ensuring compliance with legal and ethical guidelines when using such datasets.

%% file: Sections/7-Acknowledgements.tex
We thank the reviewers for their constructive feedback, which was pivotal to the addition of a few experimental results and analysis to the paper. 
This research was supported by Adobe Gift funding to the University of Houston collaboration and by the National Science Foundation (NSF) under grant \#1910192.
The authors acknowledge the use of the Sabine Cluster and the advanced support from the Research Community Data Core at the University of Houston to carry out the research presented here. 

%% file: Sections/5-Appendix.tex
\begin{table}[!h]
\centering
\resizebox{\linewidth}{!}{%
\begin{tabular}{|l|r|rrrrr|}
\hline
\multicolumn{1}{|c|}{\multirow{2}{*}{\textbf{Domain}}} &
  \multicolumn{1}{c|}{\textbf{}} &
  \multicolumn{5}{c|}{\textbf{Aspect Terms}} \\ \cline{2-7} 
\multicolumn{1}{|c|}{} &
  \multicolumn{1}{c|}{\textbf{\#Sent}} &
  \multicolumn{1}{c|}{\textbf{Pos}} &
  \multicolumn{1}{c|}{\textbf{Neg}} &
  \multicolumn{1}{c|}{\textbf{Neu}} &
  \multicolumn{1}{c|}{\textbf{Con}} &
  \multicolumn{1}{c|}{\textbf{Total}} \\ \hline
Laptops &
  348 &
  \multicolumn{1}{r|}{185} &
  \multicolumn{1}{r|}{33} &
  \multicolumn{1}{r|}{169} &
  \multicolumn{1}{r|}{1} &
  388 \\ \hline
Mobiles &
  1141 &
  \multicolumn{1}{r|}{600} &
  \multicolumn{1}{r|}{210} &
  \multicolumn{1}{r|}{578} &
  \multicolumn{1}{r|}{28} &
  1416 \\ \hline
Tablets &
  1244 &
  \multicolumn{1}{r|}{418} &
  \multicolumn{1}{r|}{157} &
  \multicolumn{1}{r|}{479} &
  \multicolumn{1}{r|}{2} &
  1056 \\ \hline
Cameras &
  150 &
  \multicolumn{1}{r|}{107} &
  \multicolumn{1}{r|}{11} &
  \multicolumn{1}{r|}{64} &
  \multicolumn{1}{r|}{1} &
  183 \\ \hline
Headphones &
  43 &
  \multicolumn{1}{r|}{20} &
  \multicolumn{1}{r|}{8} &
  \multicolumn{1}{r|}{19} &
  \multicolumn{1}{r|}{0} &
  47 \\ \hline
Home appliances &
  84 &
  \multicolumn{1}{r|}{10} &
  \multicolumn{1}{r|}{0} &
  \multicolumn{1}{r|}{34} &
  \multicolumn{1}{r|}{0} &
  44 \\ \hline
Speakers &
  47 &
  \multicolumn{1}{r|}{20} &
  \multicolumn{1}{r|}{3} &
  \multicolumn{1}{r|}{25} &
  \multicolumn{1}{r|}{0} &
  48 \\ \hline
Smartwatches &
  330 &
  \multicolumn{1}{r|}{47} &
  \multicolumn{1}{r|}{22} &
  \multicolumn{1}{r|}{149} &
  \multicolumn{1}{r|}{2} &
  220 \\ \hline
Televisions &
  135 &
  \multicolumn{1}{r|}{41} &
  \multicolumn{1}{r|}{3} &
  \multicolumn{1}{r|}{99} &
  \multicolumn{1}{r|}{1} &
  144 \\ \hline
Mobile apps &
  229 &
  \multicolumn{1}{r|}{98} &
  \multicolumn{1}{r|}{20} &
  \multicolumn{1}{r|}{46} &
  \multicolumn{1}{r|}{0} &
  164 \\ \hline
Travels &
  776 &
  \multicolumn{1}{r|}{273} &
  \multicolumn{1}{r|}{19} &
  \multicolumn{1}{r|}{98} &
  \multicolumn{1}{r|}{0} &
  390 \\ \hline
Movies &
  890 &
  \multicolumn{1}{r|}{167} &
  \multicolumn{1}{r|}{83} &
  \multicolumn{1}{r|}{154} &
  \multicolumn{1}{r|}{5} &
  409 \\ \hline
Overall &
  5417 &
  \multicolumn{1}{r|}{1986} &
  \multicolumn{1}{r|}{569} &
  \multicolumn{1}{r|}{1914} &
  \multicolumn{1}{r|}{40} &
  4509 \\ \hline
\end{tabular}%
}
\caption{Hindi Multi-Domain Dataset Statistics}
\label{tab: Hindi-Multi-Domain-Stats}
\end{table}

    \begin{figure*}
    \centering
    \includegraphics[width=\textwidth]{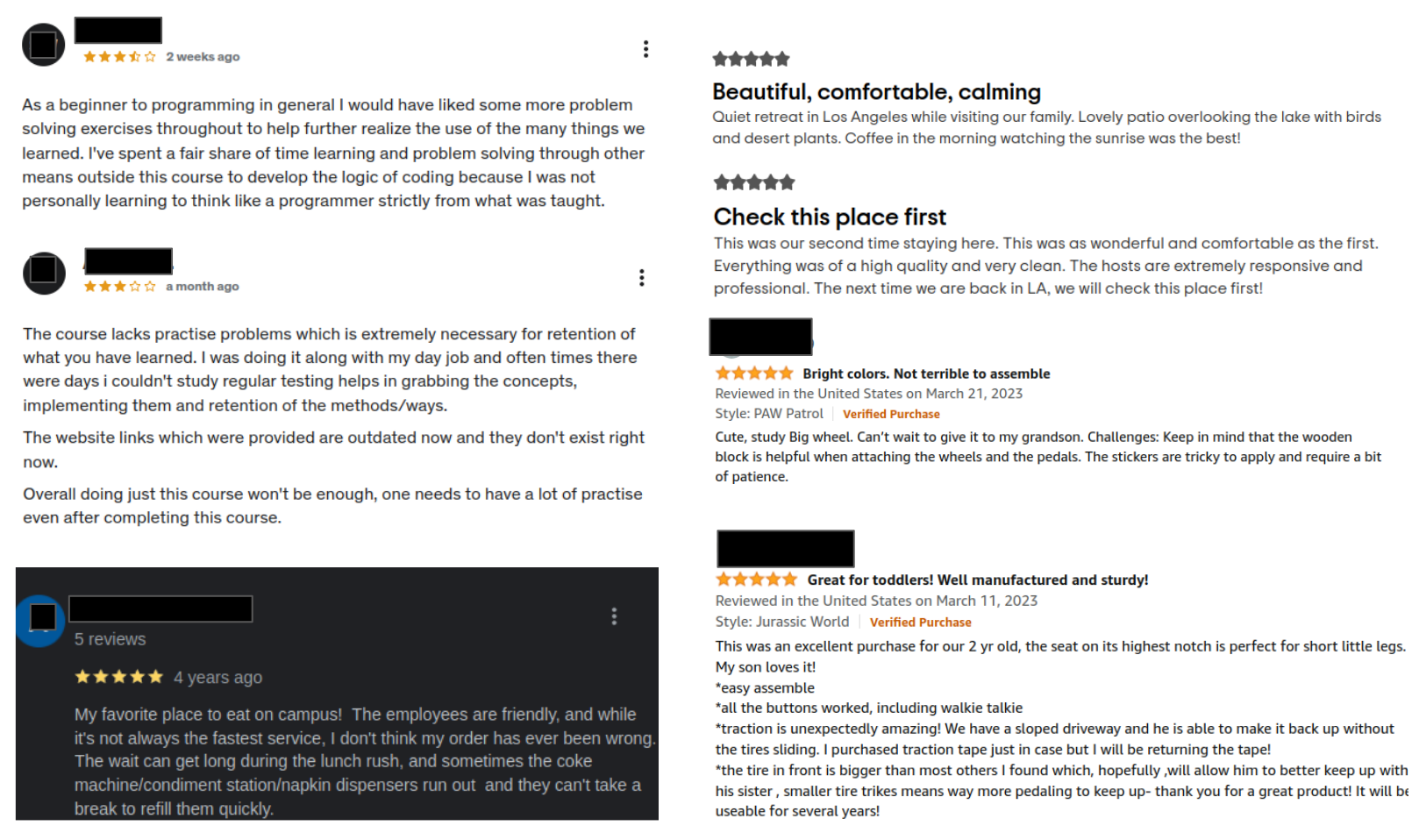}
    \caption{Multi sentence reviews from different platforms.}
    \label{fig: Multi-sent-reviews}
    \end{figure*}

    \begin{figure*}
        \centering
        \includegraphics[width=\textwidth]{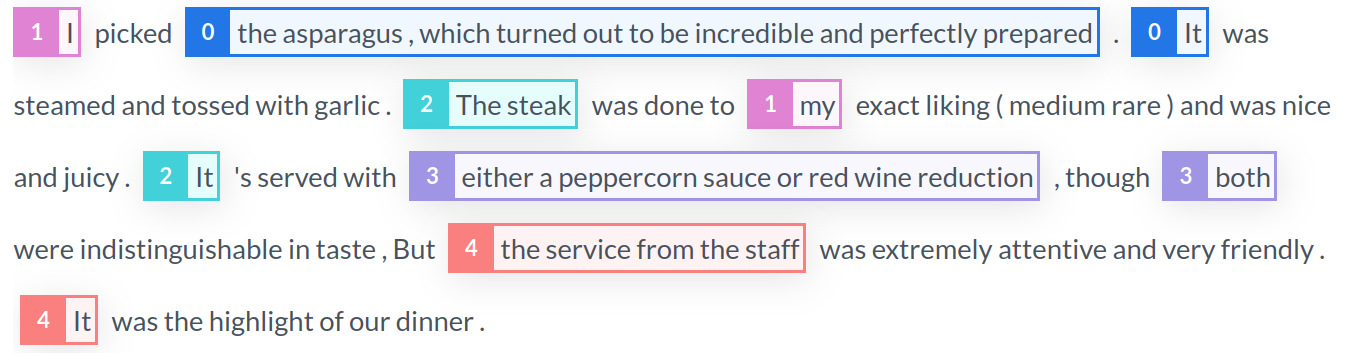}
        \caption{Example of a full restaurant review with co-references resolved in each sentence using the context from previous sentences of the review.}
        \label{fig: coref-rev}
    \end{figure*}

    \begin{table*}[!h]
        \centering
        \resizebox{\textwidth}{!}{%
        \begin{tabular}{llll}
        \hline
        \multicolumn{1}{c}{{\color[HTML]{666666} \textbf{Target}}} &
          \multicolumn{1}{c}{{\color[HTML]{666666} \textbf{Aspect Cat.}}} &
          \multicolumn{1}{c}{{\color[HTML]{666666} \textbf{Opinion Expr}}} &
          \multicolumn{1}{c}{{\color[HTML]{666666} \textbf{Sent.}}} \\ \hline
        {\color[HTML]{0000FF} } &
          {\color[HTML]{6666FF} FOOD\#QLTY} &
          {\color[HTML]{707003} incredible and perfectly prepared} &
          {\color[HTML]{CC0000} Positive} \\  
        \multirow{-2}{*}{{\color[HTML]{0000FF} asparagus}} &
          {\color[HTML]{6666FF} FOOD\#STY\_OP} &
          {\color[HTML]{707003} NULL} &
          {\color[HTML]{CC0000} Positive} \\ \hline
        {\color[HTML]{0000FF} } &
          {\color[HTML]{6666FF} FOOD\#QLTY} &
          {\color[HTML]{707003} nice and juicy} &
          {\color[HTML]{CC0000} Positive} \\  
        {\color[HTML]{0000FF} } &
          {\color[HTML]{6666FF} FOOD\#STY\_OP} &
          {\color[HTML]{707003} served with either a peppercorn sauce or red wine reduction} &
          {\color[HTML]{CC0000} Positive} \\  
        \multirow{-3}{*}{{\color[HTML]{0000FF} steak}} &
          {\color[HTML]{6666FF} FOOD\#STY\_OP} &
          {\color[HTML]{707003} indistinguishable} &
          {\color[HTML]{CC0000} Negative} \\ \hline
        {\color[HTML]{0000FF} } &
          {\color[HTML]{6666FF} } &
          {\color[HTML]{707003} extremely attentive and very friendly} &
          {\color[HTML]{CC0000} Positive} \\  
        \multirow{-2}{*}{{\color[HTML]{0000FF} service from the staff}} &
          \multirow{-2}{*}{{\color[HTML]{6666FF} SERVICE\#GEN}} &
          {\color[HTML]{707003} highlight} &
          {\color[HTML]{CC0000} Positive} \\ \hline
        \end{tabular}%
        }
        \caption{Transformation of Target, Aspect, Opinion, and Sentiment using the extended review level context, for Example in Figure \ref{fig: coref-rev}.}
        \label{tab: ex-OATS}
    \end{table*}

\subsection{Dataset Info Tables}
        
        \begin{table*}[!ht]
        \scriptsize
        \centering
        \resizebox{\textwidth}{!}{%
            \begin{tabular}{|p{4cm}|l|l|l|l|l|l|l|l|}
            \hline
            \multicolumn{1}{|l|}{\multirow{3}{*}{\textbf{\textbf{Dataset Paper}}}} &
              \multicolumn{1}{c|}{\multirow{3}{*}{\textbf{Source}}} &
              \multicolumn{1}{c|}{\multirow{3}{*}{\textbf{Domain}}} &
              \multicolumn{1}{c|}{\multirow{3}{*}{\textbf{Lng}}} &
              \multicolumn{5}{c|}{\textbf{Stats}} \\ \cline{5-9}
            \multicolumn{1}{|c|}{} &
              \multicolumn{1}{c|}{} &
              \multicolumn{1}{c|}{} &
              \multicolumn{1}{c|}{} &
             \multicolumn{1}{c|}{\multirow{2}{*}{\textbf{\#Revs}}} &
              \multicolumn{1}{c|}{\multirow{2}{*}{\textbf{\#Sent}}} &
             \multicolumn{3}{c|}{\textbf{Target/Aspect Term}} \\ \cline{7-9}
             
             \multicolumn{1}{|c|}{} &
              \multicolumn{1}{c|}{} &
              \multicolumn{1}{c|}{} &
             \multicolumn{1}{c|}{} &
              \multicolumn{1}{c|}{} &
              \multicolumn{1}{c|}{} &
             \multicolumn{1}{c|}{\#pos} &
              \multicolumn{1}{c|}{\#neg} &
             \multicolumn{1}{c|}{\#neu} \\ \hline
             
            \multicolumn{1}{|l|}{\multirow{5}{*}{Customer Reviews \cite{Bing-Liu-Products1}}} &
              \multicolumn{1}{l|}{\multirow{5}{*}{Amazon.com}}  & 
              \href{http://www.cs.uic.edu/~liub/FBS/CustomerReviewData.zip}{Cameras} &
            EN &
\multicolumn{1}{l|}{34} &
\multicolumn{1}{l|}{346} &
\multicolumn{1}{l|}{172} &
\multicolumn{1}{l|}{31} &
\multicolumn{1}{l|}{-} \\\cline{3-9}

              \multicolumn{1}{|l|}{} &
              &
              \href{http://www.cs.uic.edu/~liub/FBS/CustomerReviewData.zip}{Phones} &
            EN &
\multicolumn{1}{l|}{41} &
\multicolumn{1}{l|}{546} &
\multicolumn{1}{l|}{252} &
\multicolumn{1}{l|}{86} &
\multicolumn{1}{l|}{-} \\\cline{3-9}
              
              \multicolumn{1}{|l|}{} &
              &
              \href{http://www.cs.uic.edu/~liub/FBS/CustomerReviewData.zip}{Electronics} &
            EN &
\multicolumn{1}{l|}{95} &
\multicolumn{1}{l|}{1716} &
\multicolumn{1}{l|}{514} &
\multicolumn{1}{l|}{331} &
\multicolumn{1}{l|}{-} \\\cline{3-9}
              
              &
              &
              \href{http://www.cs.uic.edu/~liub/FBS/CustomerReviewData.zip}{DVD-player} &
            EN &
\multicolumn{1}{l|}{99} &
\multicolumn{1}{l|}{739} &
\multicolumn{1}{l|}{195} &
\multicolumn{1}{l|}{235} &
\multicolumn{1}{l|}{-} \\\cline{3-9}
              
              \multicolumn{1}{|l|}{} &
              &
              \href{http://www.cs.uic.edu/~liub/FBS/CustomerReviewData.zip}{Cameras} &
            EN &
\multicolumn{1}{l|}{45} &
\multicolumn{1}{l|}{597} &
\multicolumn{1}{l|}{224} &
\multicolumn{1}{l|}{61} &
\multicolumn{1}{l|}{-} \\\hline
            \multicolumn{1}{|l|}{\multirow{9}{*}{Customer Reviews \cite{Bing-Liu-Products2} }} &
              \multicolumn{1}{l|}{\multirow{9}{*}{Amazon.com}} &
              \href{http://www.cs.uic.edu/~liub/FBS/Reviews-9-products.rar}{Cameras} &
            EN &
\multicolumn{1}{l|}{51} &
\multicolumn{1}{l|}{300} &
\multicolumn{1}{l|}{164} &
\multicolumn{1}{l|}{58} &
\multicolumn{1}{l|}{-} \\\cline{3-9}
            \multicolumn{1}{|c|}{} & 
              &
              \href{http://www.cs.uic.edu/~liub/FBS/Reviews-9-products.rar}{Cameras} &
            EN &
\multicolumn{1}{l|}{1} &
\multicolumn{1}{l|}{229} &
\multicolumn{1}{l|}{121} &
\multicolumn{1}{l|}{27} &
\multicolumn{1}{l|}{-} \\\cline{3-9}
            \multicolumn{1}{|c|}{} & 
              &
              \href{http://www.cs.uic.edu/~liub/FBS/Reviews-9-products.rar}{Routers} &
            EN &
\multicolumn{1}{l|}{31} &
\multicolumn{1}{l|}{312} &
\multicolumn{1}{l|}{186} &
\multicolumn{1}{l|}{79} &
\multicolumn{1}{l|}{-} \\\cline{3-9}
            \multicolumn{1}{|c|}{} & 
              &
              \href{http://www.cs.uic.edu/~liub/FBS/Reviews-9-products.rar}{Phones} &
            EN &
\multicolumn{1}{l|}{49} &
\multicolumn{1}{l|}{594} &
\multicolumn{1}{l|}{310} &
\multicolumn{1}{l|}{148} &
\multicolumn{1}{l|}{-} \\\cline{3-9}
            \multicolumn{1}{|c|}{} & 
              &
              \href{http://www.cs.uic.edu/~liub/FBS/Reviews-9-products.rar}{Routers} &
            EN &
\multicolumn{1}{l|}{48} &
\multicolumn{1}{l|}{577} &
\multicolumn{1}{l|}{154} &
\multicolumn{1}{l|}{64} &
\multicolumn{1}{l|}{-} \\\cline{3-9}
             & 
              &
              \href{http://www.cs.uic.edu/~liub/FBS/Reviews-9-products.rar}{Ipod} &
            EN &
\multicolumn{1}{l|}{-} &
\multicolumn{1}{l|}{531} &
\multicolumn{1}{l|}{129} &
\multicolumn{1}{l|}{62} &
\multicolumn{1}{l|}{-} \\\cline{3-9}
            \multicolumn{1}{|c|}{} & 
              &
              \href{http://www.cs.uic.edu/~liub/FBS/Reviews-9-products.rar}{Mp3 player} &
            EN &
\multicolumn{1}{l|}{50} &
\multicolumn{1}{l|}{1011} &
\multicolumn{1}{l|}{406} &
\multicolumn{1}{l|}{177} &
\multicolumn{1}{l|}{-} \\\cline{3-9}
            \multicolumn{1}{|c|}{} & 
              &
              \href{http://www.cs.uic.edu/~liub/FBS/Reviews-9-products.rar}{Diaper Champ} &
            EN &
\multicolumn{1}{l|}{48} &
\multicolumn{1}{l|}{375} &
\multicolumn{1}{l|}{183} &
\multicolumn{1}{l|}{56} &
\multicolumn{1}{l|}{-} \\\cline{3-9}
            \multicolumn{1}{|c|}{} & 
              &
              \href{http://www.cs.uic.edu/~liub/FBS/Reviews-9-products.rar}{Antivirus} &
            EN &
\multicolumn{1}{l|}{46} &
\multicolumn{1}{l|}{380} &
\multicolumn{1}{l|}{72} &
\multicolumn{1}{l|}{169} &
\multicolumn{1}{l|}{-} \\\hline
            \multicolumn{1}{|l|}{\multirow{2}{*}{JDPA \cite{JDPA-Sentiment-Corpus}}} &
              \multicolumn{1}{l|}{\multirow{2}{*}{J.D. Power and Associates}} & 
              \href{https://verbs.colorado.edu/jdpacorpus/JDPASentimentCorpus.tar.gz}{Cameras} &
            EN &
\multicolumn{1}{l|}{96} &
\multicolumn{1}{l|}{3527} &
\multicolumn{1}{l|}{-} &
\multicolumn{1}{l|}{-} &
\multicolumn{1}{l|}{-} \\\cline{3-9}
            \multicolumn{1}{|c|}{} & 
              & 
              \href{https://verbs.colorado.edu/jdpacorpus/JDPASentimentCorpus.tar.gz}{Cars} &
            EN &
\multicolumn{1}{l|}{111} &
\multicolumn{1}{l|}{4493} &
\multicolumn{1}{l|}{-} &
\multicolumn{1}{l|}{-} &
\multicolumn{1}{l|}{-} \\\hline
            \multicolumn{1}{|l|}{\multirow{2}{*}{Darmstadt Service \cite{Darmstadt-Service-Review-Corpus} }} &
              tu-darmstadt.de & 
              \href{https://tudatalib.ulb.tu-darmstadt.de/handle/tudatalib/2448}{University reviews} &
            EN &
\multicolumn{1}{l|}{240} &
\multicolumn{1}{l|}{2786} &
\multicolumn{1}{l|}{-} &
\multicolumn{1}{l|}{-} &
\multicolumn{1}{l|}{-} \\\cline{2-9}
             & 
              Darmstadt car service & 
              \href{https://tudatalib.ulb.tu-darmstadt.de/handle/tudatalib/2448}{Service Reviews} &
            EN &
\multicolumn{1}{l|}{234} &
\multicolumn{1}{l|}{6091} &
\multicolumn{1}{l|}{-} &
\multicolumn{1}{l|}{-} &
\multicolumn{1}{l|}{-} \\\hline
               
            Twitter Comments \cite{Target-Dependent-Twitter-Sentiment} &
              www.twitter.com & 
              \href{http://goo.gl/5Enpu7}{Twitter comments} &
            EN &
\multicolumn{1}{l|}{-} &
\multicolumn{1}{l|}{6940} &
\multicolumn{1}{l|}{1734} &
\multicolumn{1}{l|}{1733} &
\multicolumn{1}{l|}{3371} \\\hline
            SE-14 \cite{14-dataset} &
              Amazon.com & 
              \href{https://alt.qcri.org/semeval2014/task4/}{Laptops} &
            EN &
\multicolumn{1}{l|}{-} &
\multicolumn{1}{l|}{3845} &
\multicolumn{1}{l|}{1331} &
\multicolumn{1}{l|}{994} &
\multicolumn{1}{l|}{629} \\\hline
            \multicolumn{1}{|l|}{\multirow{3}{*}{Customer Reviews \cite{Bing-Liu-Products3}}} &
              \multicolumn{1}{l|}{\multirow{3}{*}{Amazon.com}} & 
              \href{http://www.cs.uic.edu/~liub/FBS/CustomerReviews-3-domains.rar}{Router} &
            EN &
\multicolumn{1}{l|}{-} &
\multicolumn{1}{l|}{879} &
\multicolumn{1}{l|}{185} &
\multicolumn{1}{l|}{122} &
\multicolumn{1}{l|}{-} \\\cline{3-9}
            \multicolumn{1}{|c|}{} & 
               & 
              \href{http://www.cs.uic.edu/~liub/FBS/CustomerReviews-3-domains.rar}{Computer} &
            EN &
\multicolumn{1}{l|}{-} &
\multicolumn{1}{l|}{581} &
\multicolumn{1}{l|}{270} &
\multicolumn{1}{l|}{84} &
\multicolumn{1}{l|}{-} \\\cline{3-9}
              & 
               & 
              \href{http://www.cs.uic.edu/~liub/FBS/CustomerReviews-3-domains.rar}{Speaker} &
            EN &
\multicolumn{1}{l|}{-} &
\multicolumn{1}{l|}{689} &
\multicolumn{1}{l|}{362} &
\multicolumn{1}{l|}{78} &
\multicolumn{1}{l|}{-} \\\hline
            
             Hindi Multi-Domain \cite{hindi-dataset} &
              News-, blog-, e-com.\,sites &
              \href{https://github.com/pnisarg/ABSA}{Laptops \& 11 others}** &

            HN &
\multicolumn{1}{l|}{-} &
\multicolumn{1}{l|}{5417} &
\multicolumn{1}{l|}{1986} &
\multicolumn{1}{l|}{569} &
\multicolumn{1}{l|}{1954} \\\hline
            
            MAMS \cite{MAMS-dataset} &
              & 
              \href{https://github.com/siat-nlp/MAMS-for-ABSA/tree/master/data}{Restaurants} &

            EN &
\multicolumn{1}{l|}{-} &
\multicolumn{1}{l|}{5297} &
\multicolumn{1}{l|}{4183} &
\multicolumn{1}{l|}{3418} &
\multicolumn{1}{l|}{6253} \\\hline
            \multicolumn{1}{|l|}{\multirow{4}{*}{TOWE \cite{TOWE-dataset}}} &
              SE-14 Restaurants & 
              \href{https://github.com/NJUNLP/TOWE/blob/master/data/14res/test.tsv}{Restaurants\textsuperscript{$\pm\pm$}} &


            EN &
\multicolumn{1}{l|}{-} &
\multicolumn{1}{l|}{2127} &
\multicolumn{3}{l|}{3508 targ.- op. pairs} \\\cline{2-9}
            \multicolumn{1}{|c|}{} & 
              SE-15 Restaurants & 
              \href{https://github.com/NJUNLP/TOWE/blob/master/data/15res/test.tsv}{Restaurants\textsuperscript{$\pm\pm$}} &
            EN &
\multicolumn{1}{l|}{-} &
\multicolumn{1}{l|}{1079} &
\multicolumn{3}{l|}{1512 targ.- op. pairs} \\\cline{2-9}
            \multicolumn{1}{|c|}{} & 
              SE-16 Restaurants & 
              \href{https://github.com/NJUNLP/TOWE/blob/master/data/15res/test.tsv}{Restaurants\textsuperscript{$\pm\pm$}} &
            EN &
\multicolumn{1}{l|}{-} &
\multicolumn{1}{l|}{1408} &
\multicolumn{3}{l|}{1969 targ.- op. pairs} \\\cline{2-9}
            \multicolumn{1}{|c|}{} & 
              SE-14 Laptops & 
              \href{https://github.com/NJUNLP/TOWE/blob/master/data/15res/test.tsv}{Laptops\textsuperscript{$\pm\pm$}} &
            EN &
\multicolumn{1}{l|}{-} &
\multicolumn{1}{l|}{1501} &
\multicolumn{3}{l|}{2116 targ.- op. pairs} \\ \hline
            
\multicolumn{1}{|l|}{\multirow{4}{*}{ASTE-v1 \cite{ASTE-dataset}}} &
              SE-14 Restaurants & 
              \href{https://github.com/xuuuluuu/SemEval-Triplet-data/tree/master/ASTE-Data-V1-AAAI2020}{Restaurants\textsuperscript{$\pm\pm$}} &
            EN &
\multicolumn{1}{l|}{-} &
\multicolumn{1}{l|}{2119} &
\multicolumn{1}{l|}{3470} &
\multicolumn{1}{l|}{839} &
\multicolumn{1}{l|}{380} \\\cline{2-9}
            \multicolumn{1}{|c|}{} & 
              SE-15 Restaurants & 
              \href{https://github.com/xuuuluuu/SemEval-Triplet-data/tree/master/ASTE-Data-V1-AAAI2020}{Restaurants\textsuperscript{$\pm\pm$}} &
            EN &
\multicolumn{1}{l|}{-} &
\multicolumn{1}{l|}{1059} &
\multicolumn{1}{l|}{1586} &
\multicolumn{1}{l|}{454} &
\multicolumn{1}{l|}{66} \\\cline{2-9}
            \multicolumn{1}{|c|}{} & 
              SE-16 Restaurants & 
              \href{https://github.com/xuuuluuu/SemEval-Triplet-data/tree/master/ASTE-Data-V1-AAAI2020}{Restaurants\textsuperscript{$\pm\pm$}} &
            EN &
\multicolumn{1}{l|}{-} &
\multicolumn{1}{l|}{1372} &
\multicolumn{1}{l|}{2065} &
\multicolumn{1}{l|}{545} &
\multicolumn{1}{l|}{97} \\\cline{2-9}
            \multicolumn{1}{|c|}{} & 
              SE-14 Laptops & 
              \href{https://github.com/xuuuluuu/SemEval-Triplet-data/tree/master/ASTE-Data-V1-AAAI2020}{Laptops\textsuperscript{$\pm\pm$}} &
            EN &
\multicolumn{1}{l|}{-} &
\multicolumn{1}{l|}{1487} &
\multicolumn{1}{l|}{1664} &
\multicolumn{1}{l|}{1012} &
\multicolumn{1}{l|}{365} \\\hline            
            \multicolumn{1}{|l|}{\multirow{4}{*}{ASTE-v2 \cite{ASTE-dataset-v2}}} &
              SE-14 Restaurants & 
              \href{https://github.com/xuuuluuu/SemEval-Triplet-data/tree/master/ASTE-Data-V2-EMNLP2020}{Restaurants\textsuperscript{$\pm\pm$}} &
            EN &
\multicolumn{1}{l|}{-} &
\multicolumn{1}{l|}{2119} &
\multicolumn{1}{l|}{2769} &
\multicolumn{1}{l|}{756} &
\multicolumn{1}{l|}{286} \\\cline{2-9}
            \multicolumn{1}{|c|}{} & 
              SE-15 Restaurants & 
              \href{https://github.com/xuuuluuu/SemEval-Triplet-data/tree/master/ASTE-Data-V2-EMNLP2020}{Restaurants\textsuperscript{$\pm\pm$}} &
            EN &
\multicolumn{1}{l|}{-} &
\multicolumn{1}{l|}{1057} &
\multicolumn{1}{l|}{1285} &
\multicolumn{1}{l|}{401} &
\multicolumn{1}{l|}{61} \\\cline{2-9}
            \multicolumn{1}{|c|}{} & 
              SE-16 Restaurants & 
              \href{https://github.com/xuuuluuu/SemEval-Triplet-data/tree/master/ASTE-Data-V2-EMNLP2020}{Restaurants\textsuperscript{$\pm\pm$}} &
            EN &
\multicolumn{1}{l|}{-} &
\multicolumn{1}{l|}{1372} &
\multicolumn{1}{l|}{1674} &
\multicolumn{1}{l|}{483} &
\multicolumn{1}{l|}{90} \\\cline{2-9}
            \multicolumn{1}{|c|}{} & 
              SE-14 Laptops & 
              \href{https://github.com/xuuuluuu/SemEval-Triplet-data/tree/master/ASTE-Data-V2-EMNLP2020}{Laptops\textsuperscript{$\pm\pm$}} &
            EN &
\multicolumn{1}{l|}{-} &
\multicolumn{1}{l|}{1487} &
\multicolumn{1}{l|}{1350} &
\multicolumn{1}{l|}{774} &
\multicolumn{1}{l|}{225} \\\hline
            ABSITA-2020 \cite{ABSITA-2020}
            &
              e-Commerce platform & 
              \href{http://www.di.uniba.it/~swap/ate_absita/dataset.html}{SD cards \& 20 others\textsuperscript{$\mp\mp$}} &
            IT &
\multicolumn{1}{l|}{-} &
\multicolumn{1}{l|}{4363} &
\multicolumn{1}{l|}{7219} &
\multicolumn{1}{l|}{1577} &
\multicolumn{1}{l|}{-} \\\hline

\multicolumn{1}{|l|}{\multirow{4}{*}{ASOTE-v1 \cite{ASOTE}}} &
              SE-14 Restaurants & 
              \href{https://github.com/l294265421/ASOTE/tree/main/ASOTE-data/absa/ASOTE}{Restaurants\textsuperscript{$\pm\pm$}} &
            EN &
\multicolumn{1}{l|}{-} &
\multicolumn{1}{l|}{4828} &
\multicolumn{1}{l|}{2987} &
\multicolumn{1}{l|}{820} &
\multicolumn{1}{l|}{283} \\\cline{2-9}
            \multicolumn{1}{|c|}{} & 
              SE-15 Restaurants & 
              \href{https://github.com/l294265421/ASOTE/tree/main/ASOTE-data/absa/ASOTE}{Restaurants\textsuperscript{$\pm\pm$}} &
            EN &
\multicolumn{1}{l|}{-} &
\multicolumn{1}{l|}{1741} &
\multicolumn{1}{l|}{1304} &
\multicolumn{1}{l|}{386} &
\multicolumn{1}{l|}{80} \\\cline{2-9}
            \multicolumn{1}{|c|}{} & 
              SE-16 Restaurants & 
              \href{https://github.com/l294265421/ASOTE/tree/main/ASOTE-data/absa/ASOTE}{Restaurants\textsuperscript{$\pm\pm$}} &
            EN &
\multicolumn{1}{l|}{-} &
\multicolumn{1}{l|}{2355} &
\multicolumn{1}{l|}{1713} &
\multicolumn{1}{l|}{459} &
\multicolumn{1}{l|}{116} \\\cline{2-9}
            \multicolumn{1}{|c|}{} & 
              SE-14 Laptops & 
              \href{https://github.com/l294265421/ASOTE/tree/main/ASOTE-data/absa/ASOTE}{Laptops\textsuperscript{$\pm\pm$}} &
            EN &
\multicolumn{1}{l|}{-} &
\multicolumn{1}{l|}{3021} &
\multicolumn{1}{l|}{1396} &
\multicolumn{1}{l|}{806} &
\multicolumn{1}{l|}{213} \\\hline            
            \multicolumn{1}{|l|}{\multirow{4}{*}{ASOTE-v2 \cite{ASOTE}}} &
              SE-14 Restaurants & 
              \href{https://github.com/l294265421/ASOTE/tree/main/ASOTE-data/absa/ASOTE-v2}{Restaurants\textsuperscript{$\pm\pm$}} &
            EN &
\multicolumn{1}{l|}{-} &
\multicolumn{1}{l|}{6040} &
\multicolumn{1}{l|}{2987} &
\multicolumn{1}{l|}{820} &
\multicolumn{1}{l|}{283} \\\cline{2-9}
            \multicolumn{1}{|c|}{} & 
              SE-15 Restaurants & 
              \href{https://github.com/l294265421/ASOTE/tree/main/ASOTE-data/absa/ASOTE-v2}{Restaurants\textsuperscript{$\pm\pm$}} &
            EN &
\multicolumn{1}{l|}{-} &
\multicolumn{1}{l|}{2507} &
\multicolumn{1}{l|}{1304} &
\multicolumn{1}{l|}{386} &
\multicolumn{1}{l|}{80} \\\cline{2-9}
            \multicolumn{1}{|c|}{} & 
              SE-16 Restaurants & 
              \href{https://github.com/l294265421/ASOTE/tree/main/ASOTE-data/absa/ASOTE-v2}{Restaurants\textsuperscript{$\pm\pm$}} &
            EN &
\multicolumn{1}{l|}{-} &
\multicolumn{1}{l|}{3377} &
\multicolumn{1}{l|}{1713} &
\multicolumn{1}{l|}{459} &
\multicolumn{1}{l|}{116} \\\cline{2-9}
            \multicolumn{1}{|c|}{} & 
              SE-14 Laptops & 
              \href{https://github.com/l294265421/ASOTE/tree/main/ASOTE-data/absa/ASOTE-v2}{Laptops\textsuperscript{$\pm\pm$}} &
            EN &
\multicolumn{1}{l|}{-} &
\multicolumn{1}{l|}{4954} &
\multicolumn{1}{l|}{1392} &
\multicolumn{1}{l|}{806} &
\multicolumn{1}{l|}{213} \\\hline
            N.MTSC \cite{NewsMTSC} &
               Financial News & 
              \href{https://github.com/fhamborg/NewsMTSC}{News} &
            EN &
\multicolumn{1}{l|}{-} &
\multicolumn{1}{l|}{3021} &
\multicolumn{1}{l|}{1396} &
\multicolumn{1}{l|}{806} &
\multicolumn{1}{l|}{213} \\\hline            
            \multicolumn{1}{|l|}{\multirow{8}{*}{DM-ASTE \cite{DM-ASTE}}} &
              \multicolumn{1}{|l|}{\multirow{8}{*}{Amazon.com \cite{MEMD-ABSA-base}}} & 
              \href{https://github.com/NJUNLP/DMASTE/tree/main/dataset}{Electronics\textsuperscript{$\pm\pm$}} &
            EN &
\multicolumn{1}{l|}{-} &
\multicolumn{1}{l|}{1994} &
\multicolumn{1}{l|}{5921} &
\multicolumn{1}{l|}{1316} &
\multicolumn{1}{l|}{306} \\\cline{3-9}
            \multicolumn{1}{|c|}{} & 
              & 
              \href{https://github.com/NJUNLP/DMASTE/tree/main/dataset}{Fashion\textsuperscript{$\pm\pm$}} &
            EN &
\multicolumn{1}{l|}{-} &
\multicolumn{1}{l|}{1217} &
\multicolumn{1}{l|}{3578} &
\multicolumn{1}{l|}{918} &
\multicolumn{1}{l|}{215}\\\cline{3-9}
            \multicolumn{1}{|c|}{} & 
              & 
              \href{https://github.com/NJUNLP/DMASTE/tree/main/dataset}{Beauty\textsuperscript{$\pm\pm$}} &
            EN &
\multicolumn{1}{l|}{-} &
\multicolumn{1}{l|}{766} &
\multicolumn{1}{l|}{2576} &
\multicolumn{1}{l|}{473} &
\multicolumn{1}{l|}{117} \\\cline{3-9}
            \multicolumn{1}{|c|}{} & 
             & 
              \href{https://github.com/NJUNLP/DMASTE/tree/main/dataset}{Home\textsuperscript{$\pm\pm$}} &
            EN &
\multicolumn{1}{l|}{-} &
\multicolumn{1}{l|}{1503} &
\multicolumn{1}{l|}{4187} &
\multicolumn{1}{l|}{1091} &
\multicolumn{1}{l|}{180} \\ \cline{3-9}
            \multicolumn{1}{|l|}{} &
               & 
              \href{https://github.com/NJUNLP/DMASTE/tree/main/dataset}{Book\textsuperscript{$\pm\pm$}} &
            EN &
\multicolumn{1}{l|}{-} &
\multicolumn{1}{l|}{484} &
\multicolumn{1}{l|}{1306} &
\multicolumn{1}{l|}{232} &
\multicolumn{1}{l|}{63} \\\cline{3-9}
            \multicolumn{1}{|c|}{} & 
               & 
              \href{https://github.com/NJUNLP/DMASTE/tree/main/dataset}{Pet\textsuperscript{$\pm\pm$}} &
            EN &
\multicolumn{1}{l|}{-} &
\multicolumn{1}{l|}{507} &
\multicolumn{1}{l|}{1263} &
\multicolumn{1}{l|}{306} &
\multicolumn{1}{l|}{78}\\\cline{3-9}
            \multicolumn{1}{|c|}{} & 
              & 
              \href{https://github.com/NJUNLP/DMASTE/tree/main/dataset}{Toy\textsuperscript{$\pm\pm$}} &
            EN &
\multicolumn{1}{l|}{-} &
\multicolumn{1}{l|}{527} &
\multicolumn{1}{l|}{1622} &
\multicolumn{1}{l|}{453} &
\multicolumn{1}{l|}{70} \\\cline{3-9}
            \multicolumn{1}{|c|}{} & 
             & 
              \href{https://github.com/NJUNLP/DMASTE/tree/main/dataset}{Grocery\textsuperscript{$\pm\pm$}} &
            EN &
\multicolumn{1}{l|}{-} &
\multicolumn{1}{l|}{526} &
\multicolumn{1}{l|}{1597} &
\multicolumn{1}{l|}{285} &
\multicolumn{1}{l|}{80} \\\hline
            \multicolumn{1}{|l|}{\multirow{4}{*}{Dom-Exp-ASTE \cite{Dom-Exp-ASTE}}} &
              SE-16 Restaurants & 
              \href{https://github.com/DAMO-NLP-SG/domain-expanded-aste}{Restaurants\textsuperscript{$\pm\pm$}} &
            EN &
\multicolumn{1}{l|}{-} &
\multicolumn{1}{l|}{2942} &
\multicolumn{1}{l|}{-} &
\multicolumn{1}{l|}{-} &
\multicolumn{1}{l|}{-} \\\cline{2-9}
            \multicolumn{1}{|c|}{} & 
              SE-16 Laptops & 
              \href{https://github.com/DAMO-NLP-SG/domain-expanded-aste}{Laptops\textsuperscript{$\pm\pm$}} &
            EN &
\multicolumn{1}{l|}{-} &
\multicolumn{1}{l|}{1446} &
\multicolumn{1}{l|}{-} &
\multicolumn{1}{l|}{-} &
\multicolumn{1}{l|}{-}\\\cline{2-9}
            \multicolumn{1}{|c|}{} & 
              Hotels \cite{Dom-Exp-ASTE-base-hotels}& 
              \href{https://github.com/DAMO-NLP-SG/domain-expanded-aste}{Hotels\textsuperscript{$\pm\pm$}} &
            EN &
\multicolumn{1}{l|}{-} &
\multicolumn{1}{l|}{2136} &
\multicolumn{1}{l|}{-} &
\multicolumn{1}{l|}{-} &
\multicolumn{1}{l|}{-} \\\cline{2-9}
            \multicolumn{1}{|c|}{} & 
             Cosmetics \cite{Dom-Exp-ASTE-base-cosmetics} & 
              \href{https://github.com/DAMO-NLP-SG/domain-expanded-aste}{Cosmetics\textsuperscript{$\pm\pm$}} &
            EN &
\multicolumn{1}{l|}{-} &
\multicolumn{1}{l|}{2468} &
\multicolumn{1}{l|}{-} &
\multicolumn{1}{l|}{-} &
\multicolumn{1}{l|}{-} \\\hline

            \end{tabular}%
            
        }
        \caption{Publicly available ABSA datasets with \underline{no aspect category annotations}. Lng: Language, \#R: Number of Reviews, \#S: Number of Sentences, \#pos: Number of positive reviews, \#neg: Number of negative reviews, \#neu: Number of neutral reviews. 
        EN: English, IT: Italian, HN: Hindi. 
        N.MTSC: NewsMTSC Dataset. 
        **: Full table in Appendix Table \ref{tab: Hindi-Multi-Domain-Stats}.   
        $\pm\pm$: indicates that those datasets have the opinion phrase annotation along with other elements. 
        $\mp\mp$: Irons, Water Bottles, Action Cameras, Razors, Phones, Printer Cartridges, Coffee Capsules, Backpacks, Hair Dryers, 2 different Movies, 2 different Books, Toy Phones, Car Light bulbs, Sweatshirts, Boots, Fans, Storage Chest, Shoe Cabinets, Personal Digital Assistants, TV streaming boxes/sticks. 
        Note: The domain column has a downloadable link to each dataset.
        }
        \label{tab: Datasets-no-ac}
        \end{table*}


        \begin{table*}[!ht]
        \centering
        \resizebox{\linewidth}{!}{%
            \begin{tabular}{|l|l|l|lllllllll|}
            \hline
            \multicolumn{1}{|c|}{\multirow{3}{*}{\textbf{\textbf{Dataset Paper}}}} &
              \multicolumn{1}{c|}{\multirow{3}{*}{\textbf{Source}}} &
              \multicolumn{1}{c|}{\multirow{3}{*}{\textbf{Domain}}} &
              \multicolumn{1}{c|}{\multirow{3}{*}{\textbf{Lng}}} &
              \multicolumn{8}{c|}{\textbf{Stats}} \\ \cline{5-12}
            \multicolumn{1}{|c|}{} &
              \multicolumn{1}{c|}{} &
              \multicolumn{1}{c|}{} &
              \multicolumn{1}{c|}{} &
             \multicolumn{1}{c|}{\multirow{2}{*}{\textbf{\#Revs}}} &
              \multicolumn{1}{c|}{\multirow{2}{*}{\textbf{\#Sent}}} &
             \multicolumn{3}{c|}{\textbf{Target/Aspect Term}} &
              \multicolumn{3}{c|}{\textbf{Aspect Category}} \\ \cline{7-12}
             
             \multicolumn{1}{|c|}{} &
              \multicolumn{1}{c|}{} &
              \multicolumn{1}{c|}{} &
             \multicolumn{1}{c|}{} &
              \multicolumn{1}{c|}{} &
              \multicolumn{1}{c|}{} &
             \multicolumn{1}{c|}{\#pos} &
              \multicolumn{1}{c|}{\#neg} &
             \multicolumn{1}{c|}{\#neu} &
              \multicolumn{1}{c|}{\#pos} &
              \multicolumn{1}{c|}{\#neg} &
             \multicolumn{1}{c|}{\#neu} \\ \hline
             
            TripAdvisor Hotels \cite{TripAvisor-AmazonMp3-dataset1} &
              www.tripadvisor .com &
              \href{https://www.cs.virginia.edu/~hw5x/dataset.html}{Hotels**} &
            EN &
\multicolumn{1}{|l|}{108K} &
\multicolumn{1}{l|}{1M} &
\multicolumn{1}{l|}{-} &
\multicolumn{1}{l|}{-} &
\multicolumn{1}{l|}{-} &
\multicolumn{1}{l|}{1.63M} &
\multicolumn{1}{l|}{153K} &
\multicolumn{1}{l|}{178K} \\ \hline
            SE-14 \cite{14-dataset} &
              CSNY \cite{Ganu-CitySearch} & 
              \href{https://alt.qcri.org/semeval2014/task4/}{Restaurants} &
            EN &
\multicolumn{1}{|l|}{-} &
\multicolumn{1}{|l|}{3841} &
\multicolumn{1}{l|}{2892} &
\multicolumn{1}{l|}{1001} &
\multicolumn{1}{l|}{829} &
\multicolumn{1}{l|}{2836} &
\multicolumn{1}{l|}{998} &
\multicolumn{1}{l|}{594} \\ \hline
            \cite{CZech-ABSA} &
              www.nejezto.cz & 
              \href{https://liks.fav.zcu.cz/sentiment/}{Restaurants} &
            CZ&
\multicolumn{1}{|l|}{-} &
\multicolumn{1}{|l|}{1244} &
\multicolumn{1}{l|}{679} &
\multicolumn{1}{l|}{725} &
\multicolumn{1}{l|}{403} &
\multicolumn{1}{l|}{521} &
\multicolumn{1}{l|}{569} &
\multicolumn{1}{l|}{246} \\ \hline
            HAAD \cite{Arabic-book} &
              LABR book reviews & 
              \href{https://github.com/msmadi/HAAD}{Books} &
            AR &
\multicolumn{1}{|l|}{-} &
\multicolumn{1}{l|}{2389} &
\multicolumn{1}{l|}{1376} &
\multicolumn{1}{l|}{1287} &
\multicolumn{1}{l|}{148} &
\multicolumn{1}{l|}{721} &
\multicolumn{1}{l|}{750} &
\multicolumn{1}{l|}{14} \\ \hline
            \multicolumn{1}{|l|}{\multirow{2}{*}{SE-15 \cite{15-dataset}}} &
              CSNY \cite{Ganu-CitySearch} & 
              \href{https://alt.qcri.org/semeval2015/task12/index.php?id=data-and-tools}{Restaurants} &
            EN &
\multicolumn{1}{|l|}{350} &
\multicolumn{1}{l|}{2000} &
\multicolumn{1}{l|}{1326} &
\multicolumn{1}{l|}{496} &
\multicolumn{1}{l|}{73} &
\multicolumn{1}{l|}{1652} &
\multicolumn{1}{l|}{749} &
\multicolumn{1}{l|}{98} \\ \cline{2-12}
            \multicolumn{1}{|c|}{} & 
              Amazon.com &
              \href{https://alt.qcri.org/semeval2015/task12/index.php?id=data-and-tools}{Laptops} &
            EN &
\multicolumn{1}{|l|}{450} &
\multicolumn{1}{l|}{2500} &
\multicolumn{1}{l|}{-} &
\multicolumn{1}{l|}{-} &
\multicolumn{1}{l|}{-} &
\multicolumn{1}{l|}{1644} &
\multicolumn{1}{l|}{1094} &
\multicolumn{1}{l|}{185} \\ \hline
            \multicolumn{1}{|l|}{\multirow{11}{*}{SE-16 \cite{16-dataset}}} &
              CSNY \cite{Ganu-CitySearch} & 
              \href{https://alt.qcri.org/semeval2016/task5/index.php?id=data-and-tools}{Restaurants} &
            EN &
\multicolumn{1}{|l|}{400} &
\multicolumn{1}{l|}{2286} &
\multicolumn{1}{l|}{1817} &
\multicolumn{1}{l|}{634} &
\multicolumn{1}{l|}{106} &
\multicolumn{1}{l|}{2268} &
\multicolumn{1}{l|}{953} &
\multicolumn{1}{l|}{145} \\ \cline{2-12}
            \multicolumn{1}{|l|}{} & 
              Amazon.com &
              \href{https://alt.qcri.org/semeval2016/task5/index.php?id=data-and-tools}{Laptops} &
            EN &
\multicolumn{1}{|l|}{530} &
\multicolumn{1}{l|}{3308} &
\multicolumn{1}{l|}{-} &
\multicolumn{1}{l|}{-} &
\multicolumn{1}{l|}{-} &
\multicolumn{1}{l|}{2118} &
\multicolumn{1}{l|}{1358} &
\multicolumn{1}{l|}{236} \\ \cline{2-12}
              \multicolumn{1}{|l|}{} &
              - & 
              \href{https://alt.qcri.org/semeval2016/task5/index.php?id=data-and-tools}{Restaurants} &
              ES &
\multicolumn{1}{|l|}{-} &
\multicolumn{1}{l|}{2691} &
\multicolumn{1}{l|}{1907} &
\multicolumn{1}{l|}{672} &
\multicolumn{1}{l|}{125} &
\multicolumn{1}{l|}{2675} &
\multicolumn{1}{l|}{948} &
\multicolumn{1}{l|}{168} \\ \cline{2-12}
              \multicolumn{1}{|l|}{} &
              - & 
              \href{https://alt.qcri.org/semeval2016/task5/index.php?id=data-and-tools}{Restaurants} &
              TR &
\multicolumn{1}{|l|}{339} &
\multicolumn{1}{l|}{1248} &
\multicolumn{1}{l|}{865} &
\multicolumn{1}{l|}{555} &
\multicolumn{1}{l|}{119} &
\multicolumn{1}{l|}{924} &
\multicolumn{1}{l|}{635} &
\multicolumn{1}{l|}{135} \\ \cline{2-12}
              \multicolumn{1}{|l|}{} &
              - & 
              \href{https://alt.qcri.org/semeval2016/task5/index.php?id=data-and-tools}{Telecom} &
              TR &
\multicolumn{1}{|l|}{-} &
\multicolumn{1}{l|}{3000} &
\multicolumn{1}{l|}{-} &
\multicolumn{1}{l|}{-} &
\multicolumn{1}{l|}{-} &
\multicolumn{1}{l|}{-} &
\multicolumn{1}{l|}{-} &
\multicolumn{1}{l|}{-} \\ \cline{2-12}
              \multicolumn{1}{|l|}{} &
              - & 
              \href{https://alt.qcri.org/semeval2016/task5/index.php?id=data-and-tools}{Hotels} &
              AR &
\multicolumn{1}{|l|}{2291} &
\multicolumn{1}{l|}{6029} &
\multicolumn{1}{l|}{7213} &
\multicolumn{1}{l|}{4003} &
\multicolumn{1}{l|}{824} &
\multicolumn{1}{l|}{7705} &
\multicolumn{1}{l|}{4556} &
\multicolumn{1}{l|}{852} \\ \cline{2-12}
              \multicolumn{1}{|l|}{} &
              - & 
              \href{https://alt.qcri.org/semeval2016/task5/index.php?id=data-and-tools}{Restaurants} &
              DU &
\multicolumn{1}{|l|}{400} &
\multicolumn{1}{l|}{2286} &
\multicolumn{1}{l|}{1016} &
\multicolumn{1}{l|}{546} &
\multicolumn{1}{l|}{145} &
\multicolumn{1}{l|}{1431} &
\multicolumn{1}{l|}{857} &
\multicolumn{1}{l|}{185} \\ \cline{2-12}
              \multicolumn{1}{|l|}{} &
              - & 
              \href{https://alt.qcri.org/semeval2016/task5/index.php?id=data-and-tools}{Mobile Phones} &
              DU &
\multicolumn{1}{|l|}{270} &
\multicolumn{1}{l|}{1697} &
\multicolumn{1}{l|}{-} &
\multicolumn{1}{l|}{-} &
\multicolumn{1}{l|}{-} &
\multicolumn{1}{l|}{1454} &
\multicolumn{1}{l|}{225} &
\multicolumn{1}{l|}{110} \\ \cline{2-12}
              \multicolumn{1}{|l|}{} &
              \cite{SentiRuEval-2015} & 
              \href{https://alt.qcri.org/semeval2016/task5/index.php?id=data-and-tools}{Restaurants} &
              RU &
\multicolumn{1}{|l|}{405} &
\multicolumn{1}{l|}{4699} &
\multicolumn{1}{l|}{3139} &
\multicolumn{1}{l|}{696} &
\multicolumn{1}{l|}{313} &
\multicolumn{1}{l|}{3973} &
\multicolumn{1}{l|}{1030} &
\multicolumn{1}{l|}{379} \\ \cline{2-12}
              \multicolumn{1}{|l|}{} &
              - & 
              \href{https://alt.qcri.org/semeval2016/task5/index.php?id=data-and-tools}{Restaurants} &
              FR &
\multicolumn{1}{|l|}{455} &
\multicolumn{1}{l|}{2429} &
\multicolumn{1}{l|}{1285} &
\multicolumn{1}{l|}{1061} &
\multicolumn{1}{l|}{289} &
\multicolumn{1}{l|}{1605} &
\multicolumn{1}{l|}{1646} &
\multicolumn{1}{l|}{233} \\ \cline{2-12}
              \multicolumn{1}{|l|}{} &
              - & 
              \href{https://alt.qcri.org/semeval2016/task5/index.php?id=data-and-tools}{Mobile Phones} &
              CH &
\multicolumn{1}{|l|}{200} &
\multicolumn{1}{l|}{9500} &
\multicolumn{1}{l|}{-} &
\multicolumn{1}{l|}{-} &
\multicolumn{1}{l|}{-} &
\multicolumn{1}{l|}{1168} &
\multicolumn{1}{l|}{794} &
\multicolumn{1}{l|}{-} \\ \cline{2-12}
              \multicolumn{1}{|l|}{} &
              - & 
              \href{https://alt.qcri.org/semeval2016/task5/index.php?id=data-and-tools}{Digital Cameras} &
              CH &
\multicolumn{1}{|l|}{200} &
\multicolumn{1}{l|}{8100} &
\multicolumn{1}{l|}{-} &
\multicolumn{1}{l|}{-} &
\multicolumn{1}{l|}{-} &
\multicolumn{1}{l|}{1153} &
\multicolumn{1}{l|}{587} &
\multicolumn{1}{l|}{-} \\ \hline
            SentiHood \cite{sentihood-dataset} &
              Yahoo Question Answering &
              \href{https://github.com/uclnlp/jack/tree/master/data/sentihood}{Urban Neighborhoods} &
            EN &
\multicolumn{1}{|l|}{-} &
\multicolumn{1}{l|}{5215} &
\multicolumn{1}{l|}{-} &
\multicolumn{1}{l|}{-} &
\multicolumn{1}{l|}{-} &
\multicolumn{1}{l|}{4305} &
\multicolumn{1}{l|}{1606} &
\multicolumn{1}{l|}{-} \\ \hline
            \multicolumn{1}{|l|}{\multirow{2}{*}{Customer Response \cite{BeerAdvocate-TripAdvisor-dataset}}} &
              www.beeradvocate .com & 
              \href{https://github.com/HKUST-KnowComp/DMSC}{Beer Advocate**} &
            EN &
\multicolumn{1}{|l|}{51K} &
\multicolumn{1}{l|}{552K} &
\multicolumn{1}{l|}{-} &
\multicolumn{1}{l|}{-} &
\multicolumn{1}{l|}{-} &
\multicolumn{1}{l|}{176K} &
\multicolumn{1}{l|}{8902} &
\multicolumn{1}{l|}{64K} \\ \cline{2-12}
            \multicolumn{1}{|l|}{} & 
              www.tripadvisor .com & 
              \href{https://github.com/HKUST-KnowComp/DMSC}{Hotels**} &
            EN &
\multicolumn{1}{|l|}{29K} &
\multicolumn{1}{l|}{375K} &
\multicolumn{1}{l|}{-} &
\multicolumn{1}{l|}{-} &
\multicolumn{1}{l|}{-} &
\multicolumn{1}{l|}{120K} &
\multicolumn{1}{l|}{66K} &
\multicolumn{1}{l|}{49.1K} \\ \hline
            GermEval-2017 \cite{germevaltask2017} &
               Internet crawling with search queries & 
              \href{http://ltdata1.informatik.uni-hamburg.de/germeval2017/}{Soc.Med., blogs, news} &
            DE &
\multicolumn{1}{|l|}{-} &
\multicolumn{1}{l|}{27.8K} &
\multicolumn{1}{l|}{2802} &
\multicolumn{1}{l|}{12.5K} &
\multicolumn{1}{l|}{1459} &
\multicolumn{1}{l|}{2815} &
\multicolumn{1}{l|}{12.6K} &
\multicolumn{1}{l|}{13.9K} \\ \hline
            FiQA \cite{FiQA-dataset} &
              Financial microblogs and headlines & 
              \href{https://sites.google.com/view/fiqa/home}{Financial**} &
            EN &
\multicolumn{1}{|l|}{1303} &
\multicolumn{1}{l|}{-} &
\multicolumn{1}{l|}{774} &
\multicolumn{1}{l|}{399} &
\multicolumn{1}{l|}{-} &
\multicolumn{1}{l|}{774} &
\multicolumn{1}{l|}{399} &
\multicolumn{1}{l|}{-} \\ \hline
            \multicolumn{1}{|l|}{Bangla Rest., Cricket} &
             FB, BBC, Daily Pronthom &
              \href{https://github.com/AtikRahman/Bangla_ABSA_Datasets}{Cricket} &
              BG &
\multicolumn{1}{|l|}{-} &
\multicolumn{1}{l|}{2691} &
\multicolumn{1}{l|}{-} &
\multicolumn{1}{l|}{-} &
\multicolumn{1}{l|}{-} &
\multicolumn{1}{l|}{571} &
\multicolumn{1}{l|}{2157} &
\multicolumn{1}{l|}{266} \\ \cline{2-12}
            \multicolumn{1}{|l|}{\cite{Bangla}} & 
             SE-14 Rest \cite{14-dataset} & 
              \href{https://github.com/AtikRahman/Bangla_ABSA_Datasets}{Restaurants} &
              BG &
\multicolumn{1}{|l|}{-} &
\multicolumn{1}{l|}{1712} &
\multicolumn{1}{l|}{-} &
\multicolumn{1}{l|}{-} &
\multicolumn{1}{l|}{-} &
\multicolumn{1}{l|}{477} &
\multicolumn{1}{l|}{1226} &
\multicolumn{1}{l|}{371} \\ \hline
            ABSITA-2018 \cite{ABSITA-2018} &
              booking.com & 
              \href{http://sag.art.uniroma2.it/absita/data/}{Hotels} &
            IT &
\multicolumn{1}{|l|}{-} &
\multicolumn{1}{l|}{9285} &
\multicolumn{1}{l|}{-} &
\multicolumn{1}{l|}{-} &
\multicolumn{1}{l|}{-} &
\multicolumn{1}{l|}{6893} &
\multicolumn{1}{l|}{5288} &
\multicolumn{1}{l|}{-} \\ \hline
           Foursquare \cite{baseline-f-lex} &
               foursquare.com/ &
              \href{https://europe.naverlabs.com/Research/Natural-Language-Processing/Aspect-Based-Sentiment-Analysis-Dataset/}{Restaurants} &
            EN &
\multicolumn{1}{|l|}{-} &
\multicolumn{1}{l|}{1006} &
\multicolumn{1}{l|}{759} &
\multicolumn{1}{l|}{108} &
\multicolumn{1}{l|}{16} &
\multicolumn{1}{l|}{947} &
\multicolumn{1}{l|}{191} &
\multicolumn{1}{l|}{19} \\ \hline
            MAMS \cite{MAMS-dataset} &
              CSNY \cite{Ganu-CitySearch} & 
              \href{https://github.com/siat-nlp/MAMS-for-ABSA/tree/master/data}{Restaurants} &
            EN &
\multicolumn{1}{|l|}{-} &
\multicolumn{1}{l|}{3849} &
\multicolumn{1}{l|}{-} &
\multicolumn{1}{l|}{-} &
\multicolumn{1}{l|}{-} &
\multicolumn{1}{l|}{2415} &
\multicolumn{1}{l|}{2606} &
\multicolumn{1}{l|}{3858} \\ \hline
            Telugu Movies \cite{telugu-dataset} &
              \href{123telugu.com}{123telugu}, \href{eenadu.net}{eenadu}, \href{telugu.samayam.com}{samayam} &
              \href{http://tiny.cc/vdxugz}{Movies} &
              TE &
\multicolumn{1}{|l|}{-} &
\multicolumn{1}{l|}{5027} &
\multicolumn{1}{l|}{2480} &
\multicolumn{1}{l|}{3251} &
\multicolumn{1}{l|}{1129} &
\multicolumn{1}{l|}{2480} &
\multicolumn{1}{l|}{3251} &
\multicolumn{1}{l|}{1129} \\ \hline
            Vietnam. Smartph. \cite{vietnamese-span-detection} &
              e-commerce sites &
              \href{https://github.com/kimkim00/UIT-ViSD4SA}{Smartphones} &
              VI &
\multicolumn{1}{|l|}{-} &
\multicolumn{1}{l|}{11122} &
\multicolumn{1}{l|}{-} &
\multicolumn{1}{l|}{-} &
\multicolumn{1}{l|}{-} &
\multicolumn{1}{l|}{21.7K} &
\multicolumn{1}{l|}{11.2K} &
\multicolumn{1}{l|}{2214} \\ \hline
              ASAP \cite{chinese-rest-asap-2021} &
              O2O e-commerce platforms &
              \href{https://github.com/Meituan-Dianping/asap/tree/master/data}{Restaurants} &
              CH &
\multicolumn{1}{|l|}{46K} &
\multicolumn{1}{l|}{-} &
\multicolumn{1}{l|}{-} &
\multicolumn{1}{l|}{-} &
\multicolumn{1}{l|}{-} &
\multicolumn{1}{l|}{169K} &
\multicolumn{1}{l|}{35K} &
\multicolumn{1}{l|}{66K} \\ \hline
              \multicolumn{1}{|l|}{\multirow{2}{*}{ASQP \cite{ASQP}}} &
              SE-15 \cite{15-dataset} &
              \href{https://github.com/IsakZhang/ABSA-QUAD}{Restaurants} &
              EN &
\multicolumn{1}{|l|}{-} &
\multicolumn{1}{l|}{1580} &
\multicolumn{1}{l|}{1407} &
\multicolumn{1}{l|}{489} &
\multicolumn{1}{l|}{68} &
\multicolumn{1}{l|}{1710} &
\multicolumn{1}{l|}{701} &
\multicolumn{1}{l|}{85} \\ \cline{2-12}
               \multicolumn{1}{|l|}{} &
              SE-16 \cite{16-dataset} &
              \href{https://github.com/IsakZhang/ABSA-QUAD}{Restaurants} &
              EN &
\multicolumn{1}{|l|}{-} &
\multicolumn{1}{l|}{2124} &
\multicolumn{1}{l|}{1811} &
\multicolumn{1}{l|}{613} &
\multicolumn{1}{l|}{110} &
\multicolumn{1}{l|}{2229} &
\multicolumn{1}{l|}{877} &
\multicolumn{1}{l|}{135} \\ \hline
                \multicolumn{1}{|l|}{\multirow{2}{*}{ACOS \cite{ACOS}}} &
              SE-16 \cite{16-dataset} &
              \href{https://github.com/NUSTM/ACOS/tree/main/data}{Restaurants} &
              EN &
\multicolumn{1}{|l|}{-} &
\multicolumn{1}{l|}{2287} &
\multicolumn{1}{l|}{2742} &
\multicolumn{1}{l|}{1518} &
\multicolumn{1}{l|}{259} &
\multicolumn{1}{l|}{3578} &
\multicolumn{1}{l|}{1879} &
\multicolumn{1}{l|}{316} \\ \cline{2-12}
               \multicolumn{1}{|l|}{} &
              Amazon.com &
              \href{https://github.com/NUSTM/ACOS/tree/main/data}{Laptops} &
              EN &
\multicolumn{1}{|l|}{-} &
\multicolumn{1}{l|}{4079} &
\multicolumn{1}{l|}{2004} &
\multicolumn{1}{l|}{663} &
\multicolumn{1}{l|}{114} &
\multicolumn{1}{l|}{2503} &
\multicolumn{1}{l|}{1007} &
\multicolumn{1}{l|}{151} \\ \hline
              \multicolumn{1}{|l|}{\multirow{5}{*}{MEMD-ABSA \cite{MEMD-ABSA}}} &
              https://nijianmo.github.io/amazon/  &
              \href{}{Books} &
              EN &
\multicolumn{1}{|l|}{986} &
\multicolumn{1}{l|}{2967} &
\multicolumn{1}{l|}{-} &
\multicolumn{1}{l|}{-} &
\multicolumn{1}{l|}{-} &
\multicolumn{1}{l|}{-} &
\multicolumn{1}{l|}{-} &
\multicolumn{1}{l|}{-} \\ \cline{3-12}
               \multicolumn{1}{|l|}{} &
              \cite{MEMD-ABSA-base} &
              \href{}{Clothing} &
              EN &
\multicolumn{1}{|l|}{928} &
\multicolumn{1}{l|}{2373} &
\multicolumn{1}{l|}{-} &
\multicolumn{1}{l|}{-} &
\multicolumn{1}{l|}{-} &
\multicolumn{1}{l|}{-} &
\multicolumn{1}{l|}{-} &
\multicolumn{1}{l|}{-} \\  \cline{2-12}
                \multicolumn{1}{|l|}{} &
              https://www.yelp.com/dataset/download &
              \href{}{Restaurant} &
              EN &
\multicolumn{1}{|l|}{940} &
\multicolumn{1}{l|}{3526} &
\multicolumn{1}{l|}{-} &
\multicolumn{1}{l|}{-} &
\multicolumn{1}{l|}{-} &
\multicolumn{1}{l|}{-} &
\multicolumn{1}{l|}{-} &
\multicolumn{1}{l|}{-} \\   \cline{2-12}
              \multicolumn{1}{|l|}{} &
              http://insideairbnb.com/get-the-data/ &
              \href{}{Hotel} &
              EN &
\multicolumn{1}{|l|}{1029} &
\multicolumn{1}{l|}{5152} &
\multicolumn{1}{l|}{-} &
\multicolumn{1}{l|}{-} &
\multicolumn{1}{l|}{-} &
\multicolumn{1}{l|}{-} &
\multicolumn{1}{l|}{-} &
\multicolumn{1}{l|}{-} \\  \cline{2-12}
              \multicolumn{1}{|l|}{} &
              Amazon.com &
              \href{}{Laptops} &
              EN &
\multicolumn{1}{|l|}{-} &
\multicolumn{1}{l|}{4076} &
\multicolumn{1}{l|}{-} &
\multicolumn{1}{l|}{-} &
\multicolumn{1}{l|}{-} &
\multicolumn{1}{l|}{-} &
\multicolumn{1}{l|}{-} &
\multicolumn{1}{l|}{-} \\ 
                            \hline
              
            \end{tabular}%
        }
        \caption{Publicly available ABSA datasets \underline{with aspect category annotations} along with aspect terms/targets and polarity. Lng: Language, \#Revs: Number of Reviews, \#Sent: Number of Sentences, \#pos: Number of positive reviews, \#neg: Number of negative reviews, \#neu: Number of neutral reviews. 
        EN: English, AR: Arabic, IT: Italian, CZ: Czech, TU: Turkish, RU: Russian, FR: French, CH: Chinese, DE: German, BG: Bangla, TE: Telugu, VI: Vietnamese, ES: Spanish, DU: Dutch.
        **: indicates that the dataset has ratings converted to categorical sentiment polarities. Note: The domain column has a downloadable link to each dataset.  
        }
        \label{tab: Datasets-with-ac}
        \end{table*}


